\pdfoutput=1
\documentclass{article}

     \PassOptionsToPackage{numbers, compress}{natbib}
     \usepackage[preprint]{neurips_2020}

\usepackage[utf8]{inputenc} % allow utf-8 input
\usepackage[T1]{fontenc}    % use 8-bit T1 fonts
\usepackage{hyperref}       % hyperlinks
\usepackage{url}            % simple URL typesetting
\usepackage{booktabs}       % professional-quality tables
\usepackage{amsfonts}       % blackboard math symbols
\usepackage{nicefrac}       % compact symbols for 1/2, etc.
\usepackage{microtype}      % microtypography

\usepackage{amsmath}        % math
\usepackage{mathtools}
\usepackage{dsfont}

\usepackage{algorithm}
\usepackage[noend]{algorithmic}
\usepackage{subcaption}
\usepackage{xcolor}

\usepackage{soul}
\usepackage{scrextend}
\usepackage{wrapfig}
\usepackage{caption}
\usepackage{subcaption}
\usepackage{hyperref}

\usepackage{amsmath}
\DeclareMathOperator{\MMD}{MMD}
\DeclareMathOperator*{\argmax}{argmax}
\DeclareMathOperator*{\argmin}{argmin}
\DeclareMathOperator{\RL}{RL}
\DeclareMathOperator{\IRL}{IRL}
\DeclareMathOperator{\Geom}{Geom}
\DeclareMathOperator{\Poisson}{Poisson}
\DeclareMathOperator{\GIRL}{GIRL}
\DeclareMathOperator{\EMMA}{EMMA}
\DeclareMathOperator{\WIEM}{WIEM}
\DeclareMathOperator{\MEGAN}{MEGAN}
\DeclareMathOperator{\GAIL}{GAIL}

\newtheorem{proposition}{Proposition}
\newtheorem{corollary}{Corollary}[proposition]

\title{A Generalised Inverse Reinforcement Learning Framework}

\author{%
  Firas JARBOUI \\
  ANEO \\
  Centre Borelli - ENS Paris-saclay  \\
  \texttt{firasjarboui@gmail.com} \\
  \And
  Vianey PERCHET \\
  Criteo AI Lab \\
  Crest, ENSAE \\
  \texttt{vianney.perchet@normalesup.org} \\
}

\begin{document}
\maketitle
\begin{abstract}
    The global objective of inverse Reinforcement Learning (IRL) is to estimate the unknown cost function of some MDP based on observed trajectories generated by (approximate) optimal policies. The classical approach consists in tuning this cost function so that associated optimal trajectories (that minimise the cumulative discounted cost, i.e. the classical RL loss) are “similar” to the observed ones. Prior contributions focused on penalising degenerate solutions and improving algorithmic scalability. Quite orthogonally to them,  we question the pertinence of characterising optimality with respect to the cumulative discounted cost as it induces an implicit bias against policies with longer mixing times. 
    State of the art value based RL algorithms circumvent this issue by solving for the fixed point of the Bellman optimality operator, a stronger criterion that is not well defined for the inverse problem.
    To alleviate this bias in IRL, we introduce an alternative training loss that puts more weights on future states which yields a reformulation of the (maximum entropy) IRL problem.
    The algorithms we devised exhibit enhanced performances (and similar tractability) than off-the-shelf ones in multiple OpenAI gym environments.
\end{abstract}

%%%%%%%%%%%%%%%%%%%%%%%%%%%%%%%%%%%%%%%%%%%%%%%%%%%%%%%%%%%%%%%%%%%%%%%%%%%%%%%%%%%%%%%%%%%%%%%%%%%%%%%%%
%% start of main body of paper

\section{Introduction}
    Modelling the behaviours of rational agents is a long active research topic. From early attempts to decompose human and animal locomotion \cite{muybridge} to more recent approaches to simulate human movements \cite{li_Todorov, mombaur, schultz}, the common thread is an underlying assumption that the  agents are acting according to some stationary policies. To rationalise these behaviours, it is natural to  assume that they are optimal with respect to some objective function, with  evidences in the case of animal conditioned learning \cite{schmajuk1997, verschure2003, maia2010, verschure2014}. 
    
    The global objective of Inverse Reinforcement Learning (IRL) is inferring such objective function given measurements of the rational agent's behaviour, its sensory inputs and a model of the environment \cite{russell1998}.  IRL  builds upon the standard Reinforcement Learning (RL)  formulation, where the goal is to find the policy that minimises discounted cumulative costs of some Markov Decision Process~\cite{puterman2014markov}. 
    It aims at finding  cost functions for which the observed behaviour is  ``approximately optimal''. However this simplistic formulation admits degenerate solutions \cite{apprenticeship}. This led to a series of innovative reformulations to lift this indeterminacy by favouring costs for which the observed behaviour is particularly better than alternative ones, namely maximum margin IRL \cite{ratliff2006} and maximum entropy IRL \cite{MEIRL1, MEIRL2}. The latter formulation ended up as the building block of recent breakthroughs, with both  tractable and highly performing algorithms \cite{finn2016, gail, airl}. These improvements provided the ground for multiple practical real-life applications \cite{MEIRL1, bougrain2012, sharifzadeh2016, jarboui2019, martinez2020}.
    %However, despite the undeniable progress of IRL, the derived algorithms still struggle in multi-task settings:  previously developed multi-task IRL algorithms benchmark their performances on environments with discrete task spaces~\cite{dimitrakakis2011, gleave2018, mendez2018, sGAIL}, or with discrete states/actions~\cite{fu2019, zhou2020}. To the extent of our knowledge, the only exception to this statement is the Meta-IRL setting with reported performances on fully continuous environment \cite{yu2019}. However, solutions to the Meta-IRL problem require additional training with task specific demonstrations before being used for the said task.
    
    We  propose an orthogonal improvement to this literature. We question the very pertinence of characterising optimality w.r.t.\ the cumulative discounted costs as it induces a bias against policies with longer mixing times. We propose an extension of this criterion  to alleviate this issue. From this novel objective, we derive reformulations for both the RL and IRL problems. We discuss the ability of existing RL algorithms to solve this new formulation and we generalise existing IRL algorithms to solve the problem under the new criterion. We back up our proposition with empirical evidence of improved performances in multiple OpenAI gym environments.
\section{Generalised optimality criterion}
In this section, we introduce the classical settings of RL and IRL, as well as the new generalised  settings we introduce to alleviate some inherent biases of current methods. 

\subsection{A classical RL Setting}
    Consider an infinite horizon Markov Decision Process (MDP) $\mathcal{M}=\{ \mathcal{S}, \mathcal{A}, \mathcal{P}, c, \gamma, p_0 \}$, where:
    \vspace{-0.3cm}
    \begin{description}
        \item[--] $\mathcal{S}$ is either a finite or a compact subset of $\mathds{R}^d$, for some dimension $d\in \mathds{N}$
        \item[--] $\mathcal{A}$ is either a finite or a compact subset of $\mathds{R}^{d'}$, for $d' \in \mathds{N}$
        \item[--] $\mathcal{P}$ is the state transition kernel, i.e., a continuous mapping from $\mathcal{S} \times \mathcal{A}$ to $\Delta(\mathcal{S})$, where $\Delta(\cdot)$ denotes the set of probability measures\footnote{The $\sigma$-field is always the Borel one.} over some set,
        \item[--] $c:\mathcal{S}\times \mathcal{A} \to \mathbb{R}$ is a continuous non-negative cost function,
        \item[--] $p_0 \in \Delta(\mathcal{S})$ is the initial state distribution, and $\gamma \in (0,1)$ is the discount factor.
    \end{description} 
    A policy $\pi$ is a mapping indicating, at each time step $t \in \mathds{N}$, the action $a_t$ to be chosen at the current state  $s_t$; it could depend on the whole past history of states/actions/rewards but it is well known that one can focus solely, at least under mild assumptions,  on stationary  policies $\pi : \mathcal{S} \to \Delta(\mathcal{A})$. The choice of a policy $\pi$, along with a kernel $\mathcal{P}$ and  the initial probability $p_0$, generates a unique probability distribution  over the sequences of states denoted by $\mathbb{P}_\pi$ (the solution to the forward Chapman–Kolmogorov equation). The expected cumulative discounted cost of this policy, in the MDP $\mathcal{M}$ is consequently equal to $\mathbb{E}_{p_0,\pi}[\sum_t \gamma^t c(s_t,a_t)] = \int_{s_0} p_0(s_0)  \sum_{t=0}^\infty  
          \int_{s_t, a_t}\hspace{-0.5cm} \gamma^t \mathbb{P}_\pi(s_t,a_t|s_0)c(s_t,a_t).$
          
    Optimal policies are minimisers of this quantity (existence is ensured  under mild assumptions \cite{puterman2014markov}).
    A standard way to compute optimal policies, is to minimise the state-action value mapping defined as: $Q^c_\pi(s,a) = c(s,a) + \sum_{t=1}^\infty \int_{s_t, a_t} \hspace{-0.5cm}\gamma^t \mathbb{P}_\pi(s_t,a_t|s)c(s_t,a_t)$.
    Indeed, the expected cumulative discounted cost of a policy is the expectation of $Q$-function against $p_0$:
    \begin{align*}
        \mathbb{E}_{p_0,\pi}[\sum_{t=0}^\infty\gamma^t c(s_t,a_t)] = \int_{s_0,a_0}\hspace{-0.5cm}  p_0(s_0) \pi(a_0|s_0)Q^c_\pi(s_0,a_0)
    \end{align*}
    
    %Inverse RL: techniques, pitfalls and generalised criteria
    \subsection{A built-in bias in the IRL formulation}
    The problem gets more complicated in Inverse Reinforcement Learning where the objective is to learn an unknown cost function $c$ whose associated optimal policy coincides with a given one $\pi_E$ (referred to as the ``expert'' policy). This problem is unfortunately ill-posed as all policies are optimal w.r.t.\ a constant cost function \cite{apprenticeship}. 
    In order to lift this indeterminacy, the most used alternative formulation is called \textsl{maximum entropy inverse reinforcement learning} \cite{MEIRL1, MEIRL2} that aims at finding a cost function $c^*$ such that the expert policy $\pi_E$ has a relatively small cumulative cost $\mathbb{E}_{p_0,\pi_E}[\sum_{t=0}^\infty\gamma^t c^*]$ while other policies incur a much higher cost. This implicitly boils down to learning an optimal policy (associated to some learned cost) that matches the expert's future state occupancy measure $\rho_{\pi_E}$ marginalised over the initial state distribution, where $\rho_\pi(s,a|s_0) = \sum_t \gamma^t \mathbb{P}_\pi(s_t = s, a_t=a|s_0)$.
    
    State of the art approaches \cite{gail,airl} consist, roughly speaking, in a two-step procedure. In the first step, given a cost function $\hat{c}$, an (approximately) optimal policy $\hat{\pi}$ of $\hat{\mathcal{M}}$ (the MDP $\mathcal{M}$ with $\hat{c}$ for  cost function), is learned. In the second step,  trajectories  generated by $\hat{\pi}$   are \textsl{compared} to expert ones (in the sense of $\rho_\pi$); then $\hat{c}$ is updated to penalise states unvisited by the expert (say, by gradient descent over some parameters). Obviously, those two steps can be repeated until convergence (or until the generated and the original data-sets are close enough).

    However,  the presence of a discount factor in the definition of $\rho_\pi$ has a huge undesirable effect: the total weight of the states in the far future (say, after some stage $t^*$) is negligible in the global loss, as it would be of the order of $\gamma^{t^*}$. So trying to match the future state  occupancy measure will implicitly favours policies  mimicking the behaviour in the short term.
    As a consequence, this would end up in penalising  policies with longer mixing times even if their stationary distribution matches the experts on the long run. 
    This built-in bias is a consequence of solving the reinforcement learning step with policies that optimise the cumulative discounted costs (minimises the expectation of the Q-functions against $p_0$) rather than policies that achieve the Bellman optimality criterion (minimises the Q-function for any state action pairs). Unfortunately, there is no IRL framework  solving the problem under the latter assumption.
    
    In order to bridge this gap, we introduce a more general optimality criterion for the reinforcement learning step; it is still defined as the expectation of the $Q$-function, yet not against $p_0$ as in traditional RL, but against \textbf{both} the initial \textbf{and} the future states distributions. To get some flexibility, we allow the loss to weight present and future states differently by considering a probability distribution $\eta$ over~$\mathbb{N}$. Formally, we define the $\eta$-weighted future state measurement distribution:
    \begin{align*}
        P^{\eta}_{\pi}(s_{+}, a_{+}|s_0):= \sum_{n=0}^\infty \eta(n) \mathbb{P}_\pi(s_n = s_{+}, a_n=a_{+}|s_0).
    \end{align*}
    Using $P_{\pi}^{\eta}$, the new criterion is defined as:
    \begin{align*}
        \mathds{E}^{\eta}_{p_0,\pi}[Q^c_{\pi}] 
        :=\int_{s_0}\hspace{-0.1cm} p_0(s_0)
             \int_{s_+,a_+}\hspace{-0.5cm} P^{\eta}_{\pi}(s_{+}, a_{+}|s_0) Q^c_\pi(s_+,a_+)  
        = \mathds{E}_{p_0,\pi}\Big[\sum_k \eta(k) \sum_t \gamma^t c_{t+k} \Big]
    \end{align*}
    where $c_t$ denotes the cost at the $t^\text{th}$ observation ($c(s_t,a_t)$).
     Any policy that minimises $\mathds{E}^{\eta}_{p_0,\pi}[Q^c_{\pi}]$ will now be referred to as ``$\eta$-optimal'' (w.r.t.\ the cost function $c$).
    As mentioned before, the inverse RL problem can be decomposed in two sub-problems, learning approximate optimal strategies (given a candidate $\hat{c}$) and optimizing over $\hat{c}$ (taking into account the expert distribution $\pi_E$). In order to avoid over-fitting when learning optimal policies, the standard way is to regularize the optimization loss \cite{regMDP}. As a consequence, we consider any mapping $\Omega: \Delta(\mathcal{A})^\mathcal{S}\rightarrow\mathbb{R}$ that is a concave over the space of policies. The associated regularised loss of adopting a policy $\pi$ given the cost function $c$ is defined as:
    \begin{equation}
        \mathcal{L}_\Omega^\eta(\pi,c) = \mathds{E}^{\eta}_{p_0,\pi}\big[ Q^{c}_{\pi} \big] - \Omega(\pi)
    \end{equation}
    The generalised RL problem is then defined as:
    \begin{equation}
        \RL^\eta_\Omega(c)  := \argmin_{\pi} \mathcal{L}_\Omega^\eta(\pi,c)
        \label{GRL_prob_formulation}
    \end{equation}
   Similarly, in order to learn simpler cost functions \cite{gail}, the optimization loss  considered is in turn penalised by  a convex (over the space of cost functions)  regularizer $\psi: \mathbb{R}^{(\mathcal{S}\times\mathcal{A})}\rightarrow\mathbb{R}$. The problem of Generalised (Maximum Entropy) Inverse Reinforcement learning, whose objective is to learn an appropriate cost function $c$, is formally defined as :
      \begin{equation} 
        \IRL^\eta_{\psi,\Omega}(\pi_E)
        := \argmax_c  \min_{\pi} \mathcal{L}_\Omega^\eta(\pi,c) -  \mathcal{L}_\Omega^\eta(\pi_E,c)   - \psi(c) 
            \label{GIRL_prob_formulation}
    \end{equation}
    We emphasise that simply choosing $\delta_0$ (a Dirac mass at 0) for the distribution $\eta$ induces the  classical definitions of both the $\RL$ and $\IRL$ problems \cite{gail}. On the other hand, choosing $\eta=\Geom(\gamma)$ transforms the loss into the expectation of the sum of discounted Q-functions along the trajectory. 
    
    Hypothetically, there could be other generalisations of discounted cost. However, preserving the compatibility of the Bellman criterion with the proposed generalisation for RL and duality properties for IRL is not trivial (for example,  polynomial decay $\frac{\gamma}{t^n}$ would break these properties). In the following, we prove that the $\eta$-optimality framework satisfies both properties.

    \subsection{Generalised Reinforcement Learning }
    \label{sec:GRL}

        As in the classical setting, solving the generalised IRL problem (Equation \ref{GIRL_prob_formulation}), requires solving the generalised RL problem (Equation \ref{GRL_prob_formulation}) as a sub-routine.
        Among the model free RL algorithms,  value-based vs.\ policy gradient-based methods can be distinguished.  In this section, we focus on the first type of methods as they  can easily be used for the search of $\eta$-optimal policies. We provide  a detailed discussion of the limitations of current policy gradient-based methods in Appendix~\ref{A:pol_grad}, as they might be less adapted to solving $\RL_\Omega^\eta$.
        
        Given a standard MDP $\mathcal{M}$ and policy $\pi$, the Bellman operator $T_\pi$ (from $\mathds{R}^{\mathcal{S}}$ to $\mathds{R}^{\mathcal{S}}$) is defined as \cite{regMDP}:
        \begin{align*}
            [T_{\pi}(v)](s) = \mathbb{E}_{a\sim\pi}\Big[ c(s,a) + \gamma \mathbb{E}_{s'|s,a}[v(s')]\Big],
        \end{align*}
        and its unique fixed point  is called the associated value function $v_\pi^c$. This concept is transposed to the regularised case as follows: Given a concave regularisation function $\Omega$ and a policy $\pi$, the associated regularised Bellman operator $T_{\pi,\Omega}$ and the associated value function  $v_{\pi,\Omega}^c$ are respectively defined as: 
        \begin{align*}
            T_{\pi,\Omega} : v \in \mathbb{R}^\mathcal{S} \rightarrow T_{\pi,\Omega}(v) = T_{\pi}(v) - \Omega(\pi) \in \mathbb{R}^\mathcal{S},
        \end{align*}
        and as: $v_{\pi,\Omega}^c= T_{\pi,\Omega}(v_{\pi,\Omega}^c)$, its unique fixed point. 
        As usual, the regularised Bellman \textsl{optimality} operator $T_{*,\Omega}$ is in turn  defined as:
        \begin{align*}
             T_{*,\Omega} : v \in \mathbb{R}^\mathcal{S} & \mapsto T_{*,\Omega}(v) \in \mathbb{R}^\mathcal{S}  \\
             [T_{*,\Omega}(v)](s) & = \min_\pi [T_{\pi,\Omega}(v)](s), \quad \forall s\in\mathcal{S}.
        \end{align*}
        Notice that given $v \in \mathds{R}^{\mathcal{S}}$, the policy $\bar{\pi}_v(\cdot|s)=\delta_{\bar{a}}$ with $\bar{a} = \argmin_a c(s,a) + \gamma \mathbb{E}_{s'|s,a}[v(s')]$ achieves the minimum in the overall equation for all state $s\in \mathcal{S}$.
        The policy improvement theorem~\cite{sutton} guarantees that if $v_{\pi,\Omega}^c$ is the regularised value function of  $\pi$, then $\bar{\pi}:=\bar{\pi}_{v_{\pi,\Omega}^c}$ dominates $\pi$ (in the sense that $v_{\bar{\pi}, \Omega}^c(s) \leq v_{\pi, \Omega}^c(s)$ for any state $s\in\mathcal{S}$).
        
        If we denote by $v_{*,\Omega}^c$ the unique fixed point of the regularised Bellman optimality operator $T_{*,\Omega}$, then the policy $\pi^*_\Omega:=\bar{\pi}_{v_{*,\Omega}^c}$ is associated to the minimum regularised value function:
        
        \begin{proposition}
        \label{PR:opt_reg_policy}
            \textbf{Optimal regularised policy (Theorem 1 of \cite{regMDP})} : The policy $\pi_\Omega^* :=\bar{\pi}_{v_{*,\Omega}^c}$ is the unique optimal regularised policy in the sense that, for all policies $\pi$, the following holds: 
            $$\forall s\in\mathcal{S}, \quad v^{c}_{\pi_\Omega^*, \Omega}(s) = v_{*,\Omega}^c(s) \leq v^c_{\pi, \Omega}(s).$$
        \end{proposition}
        Quite interestingly, the  optimal regularised policy $\pi_\Omega^*$ (that minimizes the regularised cumulative discounted cost), is still minimizing the regularised $\eta$-weighted Q-functions:
        \begin{corollary}
            For any given distribution $\eta$, the optimal regularised policy minimises $\RL_\Omega^\eta$: 
            $$ \pi_\Omega^* \in \argmin_{\pi} \mathds{E}^{\eta}_{p_0,\pi}\big[ Q^c_{\pi} \big] - \Omega(\pi) $$ 
            \label{C:gen_opt_reg_policy}
        \end{corollary}
        This implies that such policy is not only optimal in the sense of the classical formulation (i.e. with $\eta = \delta_0$), but also $\eta$-optimal for any given distribution $\eta$.
        %This implies that solving the generalised $\RL_\Omega^\eta$ formulation is `equivalent' to solving the classical one (i.e. with $\eta = \delta_0$). 
        More importantly, we can directly exploit state of the art value base RL algorithms that approximates $v_{*,\Omega}$ (such as Soft Actor Critic (SAC) \cite{sac}) to solve the generalised setting. Corollary \ref{C:gen_opt_reg_policy} guarantees that the Bellman optimality criterion is compatible with the proposed generalisation ($\RL_\Omega^\eta$) for any distribution $\eta$.

    \subsection{Generalised Inverse Reinforcement Learning}
    \label{sec:GIRL}
        We recall that the global objective of IRL is to learn the cost function based on an expert policy $\pi_E$. In this section, we illustrate that the solution of $\IRL_{\psi,\Omega}^\eta(\pi_E)$ is a cost function~$\hat{c}$, whose associated optimal policy $\RL_\Omega^\eta(\hat{c})$ matches the expert's future state distributions $P^\eta_{\pi_E}$ marginalised against $\rho_{\pi_E}$ rather than simply matching the occupancy measure $\rho_{\pi_E}$, as in usual IRL formulation.
        To alleviate notations, we denote the $\eta$-optimal policy $\hat{\pi}=\RL_\Omega^\eta(\hat{c})$ as $\RL_\Omega^\eta \circ \IRL_{\psi,\Omega}^\eta (\pi_E)$.
        This policy minimises the worst-case cost weighted divergence $d_c(\hat{\pi}\|\pi_E): \mathcal{S}\mapsto\mathbb{R}$ averaged over $p_0$, such that:
        \begin{align*}
            d_c(\hat{\pi}\|\pi_E)(s_0)
            := \hspace{-0.2cm} \int_{s,a,s_+,a_+} \hspace{-1.1cm} c(s_+,a_+) \Big[  \rho_{\hat{\pi}}(s,a|s_0)P_{\hat{\pi}}^{\eta}(s_{+}, a_{+}|s,a) -\rho_{\pi_E}(s,a|s_0)P_{\pi_E}^{\eta}(s_{+}, a_{+}|s,a) \Big]
        \end{align*}
        This is formalised in the following proposition that requires the following notations. Given a policy $\pi$ we denote by $\mu_\pi(s_+,a_+|s_0) = \sum_{t,k} \gamma^t \eta(k) \mathbb{P}_\pi(s_{t+k}=s_+,a_{t+k}=a_+|s_0)$ the frequency  of $(s_{+},a_{+})$ in the $\eta$-weighted future steps of trajectories initialised according to $\rho_\pi(s,a|s_0)$.
        \begin{proposition}\label{PR:equivalent}
            For any convex penalty $\psi$, concave regulariser $\Omega$ (w.r.t.\ the future occupancy measure $\mu_\pi$) and any expert policy $\pi_E$, if  $\eta$ is geometric, then:
            \begin{align*}
                &\RL_\Omega^\eta \circ \IRL_{\psi,\Omega}^\eta (\pi_E) = \argmin_\pi \max_c L(\pi,c) \\
                \text{where: } \;& L(\pi, c) = -\Omega(\pi) - \psi(c) + \int_{s_0} p_0(s_0) d_c(\pi\|\pi_E)(s_0)
            \end{align*}
        \end{proposition}
    %{\color{red} Ici, il faudrait un statement du genre : Si $\Omega$ est concave wrt something alors $\Omega$ est concave wrt $\mu_\pi$ }{\color{blue} Pending}
    
        As mentioned before, Proposition \ref{PR:equivalent} states that  in its generalised formulation, solving the IRL problem can be done  by matching $\eta$-weighted future state distributions $\mu_\pi$ (as opposed to matching $\rho_\pi$ in the classical case).
        This proves that the generalised setting preserves the duality properties of classical IRL.
        The solution of $\IRL_{\psi, \Omega}^\eta$ is a \hbox{Nash-Equilibrium} of a game between poicies and the cost~functions:  
        \begin{corollary}\label{CR:Nash}
            Under the assumptions of Proposition \ref{PR:equivalent},  $(\Tilde{c},\Tilde{\pi})=  (\IRL_{\psi,\Omega}^\eta(\pi_E),  \RL_\Omega^\eta(\Tilde{c}))$ is a Nash-Equilibrium of the following game: 
            \begin{equation*}
                \begin{array}{ll}
                    \Tilde{c}: \max_c L(\pi,c) + \Omega(\pi) \quad ; \quad
                    \Tilde{\pi}: \min_\pi \mathds{E}_+^\pi\big[ Q_{\pi}(c) \big] - \Omega(\pi)
                \end{array}
            \end{equation*}
        \end{corollary}
        
        A practical implication of  Corollary \ref{CR:Nash} is a template algorithm  dubbed $\GIRL_{(\psi,\Omega,\eta)}$ and illustrated in Algorithm \ref{GIRL} that can be used to solve this problem.
        
        \begin{algorithm}
            \caption{$\GIRL_{(\psi,\Omega,\eta)}$ (Generalised IRL)}\label{GIRL}
            \begin{algorithmic}[1]
                \STATE {\bfseries Input:} Expert trajectories $\tau_E \sim \pi_E$, initial policy $\pi_{\theta_0}$ and initial cost function $c_{w_0}$
                \FOR{$e \in [1, N]$}
                \STATE Sample trajectories $\tau \sim \pi_{\theta_i}$
                \STATE Sample from $\tau$ policy state action $(S^+,A^+)\sim \mu_{\pi}^{\eta}(\tau)$
                \STATE Sample from $\tau_E$ expert state action $(S^+_E,A^+_E)\sim \mu_{\pi_E}^{\eta}(\tau_E)$
                \STATE Update the cost parameter $w_i$ to maximise \hbox{$-\psi(c_w) + \sum_{S^+,A^+}c_{w}(s,a) - \sum_{S^+_E,A^+_E}c_{w}(s,a)$}
                 \STATE Update $\theta_i$ using a value-based reinforcement learning algorithm to minimise $c_{w_{i+1}}$ 
                \ENDFOR
                \STATE {\bfseries Return:} $(\pi_{\theta_N}, c_{w_N})$
            \end{algorithmic}
        \end{algorithm}

        We stress out now that the concavity of $\Omega$ w.r.t.\ $\mu_\pi$ in Proposition \ref{PR:equivalent} is not too restrictive in practical settings as the $\eta$-weighted entropy regulariser, amongst others, satisfies it:
        \begin{proposition} The $\eta$-weighted entropy regulariser
            $\bar{H}^{\eta}_{p_0}$ defined by
            \begin{align*}
                \bar{H}^{\eta}_{p_0}(\mu_\pi) := H^{\eta}_{p_0}(\pi) = \mathds{E}^{\eta}_{p_0,\pi}\Big[\sum_t -\gamma^t\log\big[\pi(a_t|s_t)\big]\Big]
            \end{align*}
            is concave with respect to the occupancy measure~$\mu_\pi$.
            \label{P:convex_entropy}
        \end{proposition}
    
\section{Tractability}
    \label{sec:tractability}
    The tractability of $\GIRL_{\psi, \Omega, \eta}$ is a crucial requirement for practical implementation. In this section, both the regulariser term $\Omega(\cdot)$ and the penalty term $\psi(\cdot)$ are assumed to be tractably optimisable. For example the entropy, a widely used regulariser in the RL literature is efficiently tractable in practice. Indeed,  Soft Actor Critic \cite{sac} uses a single sample approximation of the entropy to optimise the entropy regularised Bellman optimality operator. Similarly, using an indicator penalty over a subset $\mathcal{C}$ of possible cost functions (i.e., the penalty is infinite if $c\notin\mathcal{C}$ and $0$ otherwise) is also tractable with projected gradient updates if $\mathcal{C}$ is convex \cite{apprenticeship, LPAL,MWAL}.
    As a consequence, establishing tractability of $\GIRL_{\psi, \Omega, \eta}$ reduces to finding tractable sampling schemes from $\mu_\pi$. This is equivalent to sampling sequentially from the $\eta$-weighted future state distribution and the occupancy measure as: 
    \begin{align*}
        \mu_\pi(s_+,a_+|s_0) = \int_{s,a} \rho_\pi(s,a|s_0) P_\pi^\eta(s_+,a_+|s,a) = \int_{s,a} P_\pi^\eta(s,a|s_0) \rho_\pi(s_+,a_+|s,a)
    \end{align*}
    
    Given a policy $\pi$, the simplest approach to sample from these distributions is to sample transitions from a set of $\pi$-generated trajectories, denoted by $\{(s^{(i)}_t,a^{(i)}_t)_{t\in\{1,H\}}; i\in\{1,N\}\}$, where $H$ is the horizon and $N$ is the number of trajectories.
    
    \textbf{-- For the occupancy measure $\rho_\pi$:} given a uniformly sampled index $i\sim\mathcal{U}[1,N]$ and a time sampled from truncated geometric distribution $t\sim\Geom_{[1,H]}(\gamma)$,  the associated pair of state/action  $(s_t^{(i)},a_t^{(i)})$ is an (approximate) sample from the marginal of $\rho_\pi(.|s_0)$ against $p_0$. \\
    \textbf{-- For the future state distribution $P_\pi^\eta$:} Given a state $s_{t}^{(i)}$ sampled as above, a time $k$ is sampled from a truncated $\eta_{[1,H-t]}$; the state-action $(s_{t+k}^{(i)}, a_{t+k}^{(i)})$ is an approximate sample from $P_\pi^\eta(.|s_t)$.
    
    As a consequence, the above scheme shows that sampling from $\rho_\pi$ and $P_\pi^\eta$ reduces to sampling indices from $\Geom(\gamma)$ and $\eta$, which is tractable from both the expert and the learned policies perspective. This proves that solving $\IRL_{\psi,\Omega}^\eta$ does not incur any additional computational burden.

\section{MEGAN: Maximum Entropy - Generative Adversarial Network}
\label{sec:settings}
    This section introduces a new algorithm, called $\MEGAN$, that will improve upon state of the art IRL algorithms. Recent progresses in the field propose variations of GAIL \cite{gail} in order to solve a wide variety of problems. For example, AIRL \cite{airl}  uses a particular shape for the discriminator for better transferability of the learned rewards, EAIRL \cite{EAIRL} applies empowerment regulariser to policy updates to prevent over-fitting the expert demonstration, RAIRL \cite{RAIRL} generalises AIRL for regularised MDPs (i.e. $\Omega$ is not necessarily the entropy), s-GAIL \cite{sGAIL} generalises the formulation for multi-task RL, etc.
        
        Their contributions were crucial to the progresses of IRL. However, we will actually focus on improving the core algorithm GAIL so that all the aforementioned approaches can be implemented with $\MEGAN$ instead of GAIL with improved performances.

        We considered the rather classical penalty function  \cite{gail}:
        \begin{equation*}
            \psi_{GAN}(c) = 
            \left\{
            \begin{array}{cc}
                \mathbb{E}_{p_0,\pi_E}^{\eta} [g(c(s,a))] & \textit{ if } c<0 \\
                +\infty &  \textit{ otherwise }
            \end{array}
            \right .
            \textbf{where: }  g(x) = 
            \left\{
            \begin{array}{cc}
                  -x - \log(1-e^x) &  \textit{if } x<0 \\
                  +\infty &  \textit{if } x\geq 0
            \end{array}
            \right .
        \end{equation*}
        
        As in the precedent cases studied, the generalised problem boils down to using $\mathbb{E}_\pi^\eta$:
        \begin{proposition}
            Under the assumptions of Proposition \ref{PR:equivalent}, and for $\psi=\psi_{GAN}(c)$, it holds:
            \begin{align*}
                    \RL_\Omega^\eta \circ \IRL_{\psi}^\eta (\pi_E) = \argmin_\pi \; -\Omega(\pi) +  \max_{D \in (0,1)^{\mathcal{S}\times\mathcal{A}}} \mathbb{E}_\pi^\eta[\log D]-\mathbb{E}_{\pi_E}^\eta[\log (1-D))] 
            \end{align*} 
            \label{P:MEGAN}
        \end{proposition}
       The algorithm $\MEGAN$ (Maximum Entropy - Generative Adversarial Network), is then equivalent to $\GIRL_{\psi_{GAN}, H, \Geom(\gamma)}$ and a generalisation of the corner stone in state of the art IRL \cite{gail}; its pseudo-code is given in Algorithm \ref{Algo:Megan}.
        
        \begin{algorithm}
            \caption{$\MEGAN$}\label{MEGAN}
            \begin{algorithmic}[1]
                \STATE {\bfseries Input:} Expert trajectories $\tau_E \sim \pi_E$, initial policy $\pi_{\theta_0}$ and initial discriminator function $D_{w_0}$
                \FOR{$e \in [1, N]$}
                \STATE Sample trajectories $\tau \sim \pi_{\theta_i}$
                \STATE Sample states randomly $(S_t,A_t)\sim \tau$ and $(S^+,A^+) = (S_{t+k}, A_{t+k})$ where $k\sim\eta$  
                \STATE Sample states randomly $(S'_t,A'_t)\sim \tau_E$ and $(S^+_E,A^+_E) = (S'_{t+k}, A'_{t+k})$ where $k\sim\eta$  
                \STATE Update the cost parameter $w_i$ to maximise \hbox{$\big[ \log D_{w}(S^+,A^+) - \log (1-D_{w}(S^+_E,A^+_E)) \big]$}
                 \STATE Update $\theta_i$ using soft actor critic to minimise the cost $\big[ \log D_{w_{i+1}}\big]$
                \ENDFOR
                \STATE {\bfseries Return:} $(\pi_{\theta_N}, D_{w_N})$
            \end{algorithmic}\label{Algo:Megan}
        \end{algorithm}
        \vspace{-0.3cm}
\section{Experiments}
\label{sec:experiments}
    This section is devoted to experimental evidences that $\MEGAN$ achieves state of the art performances. It is compared to GAIL as all subsequent approaches build upon its formulation.
    %The considered environments come with a ground truth cost function that we used to learn an expert behaviour using SAC. The obtained policy is then used to generate the expert trajectories which we use to solve IRL. 
    The standard approach to compare IRL algorithms is to consider the best performing policies obtained during the training and evaluate their performances. This is an issue in practice as we do not have access to such cost function in order to implement a stopping rule once the learned policy reaches a certain performance threshold. A reasonable alternative criterion is to measure the divergence between generated and expert future state distributions (in the sense of $\rho_{\pi}$ or $\mu_{\pi}$). 
    In this section, we propose to evaluate the divergence using the maximum mean discrepancy~(MMD)\footnote{A formal reminder on the definition of MMD divergence is provided in Appendix \ref{A:MMD} for completeness.}. We will tackle the following questions empirically:
    
    \begin{description}
        \item[-1] How does varying the parameter of a geometric $\eta$ distribution affect performances? 
        \item[-2] How does alternative $\eta$ distribution (e.g. a Poisson) compare to the use of a geometric one?
        \item[-3] Does varying the discount factor $\gamma$ produce similar performances?
    \end{description}
        
    Due to limited space, we only analyse single-task environments in this section. We provide in Appendix \ref{A:multi_task_setting} and \ref{A:experiments_multi-task} further investigations for the multi-task setting. A summary of the used hyper-parameters is also provided in Appendix \ref{A:Hyper_parameters}.
    
    \subsection{Performance improvement using a Geometric $\eta$ distribution}
    Recall that solving the IRL problem essentially boils down to finding an equilibrium between a policy that matches the expert behaviour and a cost function that discriminates generated trajectories from expert ones. An important property of a given algorithm is the stability of the associated equilibrium. In order to take into account this aspect, we propose to evaluate performances using trajectories sampled during the last $100$ iterations of training. We will refer to these trajectories as the remaining replay buffer. This procedure provides an evaluation of the policies toward which the algorithm converges, while factoring in their stability.
    
    Notice that the goal of $\GAIL$ is to match the distribution $\rho_{\pi_E}$ while $\MEGAN$ matches the distribution $\mu_{\pi_E}^\eta$. In order to take into account this difference, we propose to measure performances in terms of cumulative costs, $\MMD_\rho = \MMD(\rho_\pi, \rho_{\pi_E})$ and $\MMD_\mu = \MMD(\mu_\pi^{GEOM(0.99)}, \mu_{\pi_E}^{GEOM(0.99)})$. 
    
    We evaluate the performances of $\MEGAN$ using a truncated\footnote{Due to obvious computational limitations, the trajectories are finite} geometric $\eta$ distribution with different parameters (specifically $\{0, 0.25, 0.5, 0.75, 1\}$). Notice that using a geometric distribution with parameter $0$ is equivalent to using a Dirac mass at $0$ (or equivalently solving the IRL problem using $\GAIL$). Similarly, using a geometric distribution with parameter $1$ is equivalent to using a uniform $\eta$ distribution. The remaining values can be seen as an interpolation between these extremes. 
    
\begin{figure}
    \centering
    \begin{subfigure}[b]{.32\linewidth}
        \centering
        \includegraphics[width=\linewidth]{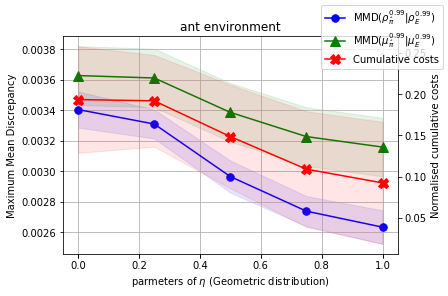}
        \caption{Ant environment}
        \label{geom_eta_perf_ant}
    \end{subfigure}
    \begin{subfigure}[b]{.32\linewidth}
        \centering
        \includegraphics[width=\linewidth]{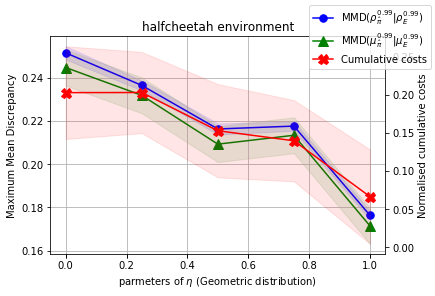}
        \caption{Half-Cheetah environment}
        \label{geom_eta_perf_half_cheetah}
    \end{subfigure}
    \begin{subfigure}[b]{.32\linewidth}
        \centering
        \includegraphics[width=\linewidth]{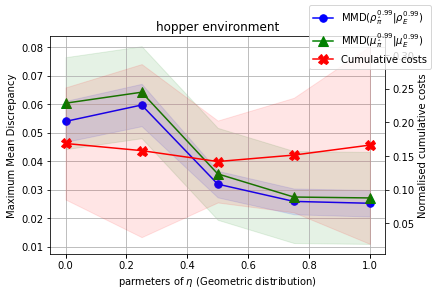}
        \caption{Hopper environment}
        \label{geom_eta_perf_hopper}
    \end{subfigure}
    \caption{ \textbf{Performances of the policies obtained during the last $100$ iteration of $\MEGAN$:} as the parameter of $\eta = \Geom(\kappa)$ grows from $0$ (equivalent to an instance of $\GAIL$) to $1$ (equivalent to an instance of $\MEGAN$ with a uniform $\eta$), we observe that the learned policies' generated trajectories are increasingly similar with those generated by the expert in the sense of $\rho_\pi$ (classical IRL criterion), $\mu_\pi$ (Generalised IRL criterion) and the cumulative discounted costs (The environment's ground truth).}
    \label{geom_eta_perf}
    \end{figure}

    In Figure \ref{geom_eta_perf}, each point reports the average performances obtained using the remaining replay buffer of $3$ randomly seeded instances of the algorithm. The blue curves report the average $\MMD_\rho$ (divergence in the sense of the classical IRL formulation), the green curves report the average $\MMD_\mu$ (divergence in the sense of the generalised IRL formulation), and the red curves report the average cumulative costs (divergence in the sense of the environment's ground truth). From left to right, we report the performances in three classical control settings with varying complexity from the \textit{MuJoCo} based environments \cite{mujoco_gym}. In Figure \ref{geom_eta_perf_ant} we used the Ant environment (a state action space of dimension $118$), in Figure \ref{geom_eta_perf_half_cheetah} we used the Half-Cheetah environment (a state action space of dimension $23$) and in Figure \ref{geom_eta_perf_hopper} we used the Hopper environment (a state action space of dimension $14$). 
    
    All the provided experiments confirmed a reduction of the average $\MMD$ divergence by $25\%$ to $60\%$ (in the sense of both classical IRL and generalised IRL formulations) as the parameter of the $\eta$ distribution increased to $1$. This confirms that using the $\eta$-optimality objective function improves both the stability and the ability of IRL algorithms to match faithfully the expert behaviour. Notice that despite the fact that $\GAIL$ explicitly optimises divergence in the sense of $\rho_\pi$, it under-performs in the sense of $\MMD_\rho$ when compared to $\MEGAN$ (the blue curves decreases as the parameter of the geometric distribution increases). This confirms empirically that the $\eta$-optimality framework proposed in this paper does indeed bridge the gap between policy-based reinforcement learning (optimising cumulative discounted costs) and value-based reinforcement learning (achieving the Bellman optimality criterion) as it even improves performances in the sense of classical IRL.
    
    Another important observation in Figure \ref{geom_eta_perf}, is that for complex environments (Ant and Half-Cheetah) the decrease of the $\MMD$ divergence -as we increased the parameter of the geometric distribution to~$1$- was correlated with a decrease of the average cumulative costs by a factor of $2$ to $4$. This was not the case of the Hopper environment, as we obtained similar cumulative costs despite the reduction of the divergence by a factor of $3$. This is explained by the fact that the ground truth cost function of the Hopper environment produces similar cumulative costs for a wider variety of policies.
    For this reason, the IRL solution does not need to achieve a faithful expert behaviour matching in order to achieve good performances. 
    This illustrates the importance of evaluating IRL algorithm with respect to $\MMD_\rho$ and $\MMD_\mu$ when the goal is to mimic behaviors. 
    
    \subsection{Performance improvement using a Poisson $\eta$ distribution}
\begin{figure}
    \centering
    \begin{subfigure}[b]{.32\linewidth}
        \centering
        \includegraphics[width=\linewidth]{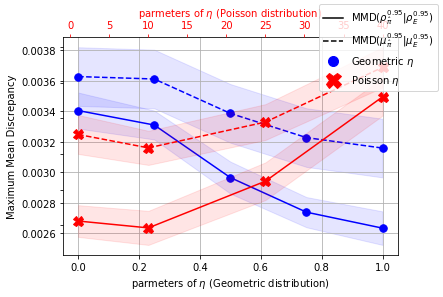}
        \caption{Ant environment}
        \label{poisson_eta_perf_ant}
    \end{subfigure}
    \begin{subfigure}[b]{.32\linewidth}
        \centering
        \includegraphics[width=\linewidth]{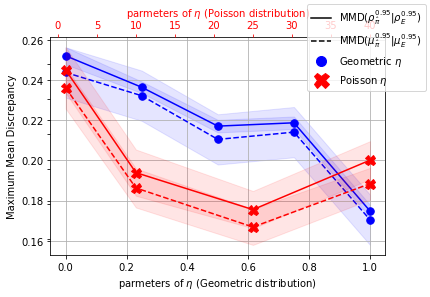}
        \caption{Half-Cheetah environment}
        \label{poisson_eta_perf_half_cheetah}
    \end{subfigure}
    \begin{subfigure}[b]{.32\linewidth}
        \centering
        \includegraphics[width=\linewidth]{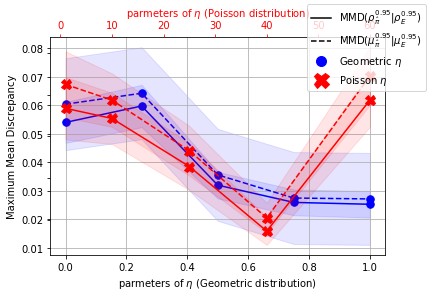}
        \caption{Hopper environment}
        \label{poisson_eta_perf_hopper}
    \end{subfigure}
    \caption{ \textbf{Performances of $\MEGAN$ using a Poisson $\eta$ distribution:} Setting the parameter of $\eta=\Poisson(\lambda)$ to a value around the length of the expert's movement cycle achieved similar/better performances than those obtained using a uniform $\eta$ distribution ($\Geom(1)$). The expert cycle is roughly $10$ frames long in the Ant environment, $25$ in the Half-Cheetah, and $40$ in the Hopper.}
    \label{poisson_eta_perf}
    \end{figure}

    Using the same experimental setting from the previous section, we evaluate $\MEGAN$'s performances with non-geometric $\eta$ distributions. 
    In Figure \ref{poisson_eta_perf}, we plot the performances of the remaining replay buffer obtained with geometric $\eta$ distributions in blue lines, and those obtained when using a Poisson distribution in red. 
    To reduce clutter, we removed the cumulative costs and only provided the divergences $\MMD_\rho$ (represented with solid lines in Figure \ref{poisson_eta_perf}) and $\MMD_\mu$ (dashed lines in Figure \ref{poisson_eta_perf}). 
    
    Despite the weaker theoretical guarantees provided in our paper for the case of non-geometric $\eta$ distributions, we observe that using a Poisson $\eta$ can lead to comparable performances. Recall that the expectation of a Poisson distribution is equal to its parameter value. This implies that solving $\IRL_{\psi, \Omega}^{Poisson(\lambda)}$ searches for policies that match $\rho_{\pi_E}(.|s)$ for states $s$ observed around the $\lambda^{\textit{th}}$ frame of the expert demonstrations. 
    Now notice that the control tasks analysed in Figure \ref{poisson_eta_perf}, consist of movements cycles that are repeated perpetually. Quite interestingly, setting $\lambda$ to a value around the length of an expert cycle ($\lambda=10$ in the Ant environment, $25$ in the Half-Cheetah, and $40$ for the Hopper), ended up achieving the best performances.
    
    In a sense, the proposed $\eta$-optimality criterion can be seen as an inductive bias: we successfully injected qualitative knowledge (the repetitive nature of the expert behaviour) by explicitly asking the agent to focus on matching $\rho_\pi(.|s)$ for states $s$ observed within a single movement cycle of the expert demonstrations via careful parameterisation of the distribution $\eta$. 
    In the case where such higher understanding/representation of the expert behaviour is unavailable, using a uniform $\eta$ distribution (or a geometric $\eta$ with a parameter close to $1$) is a safe bet. Notice that in Figure \ref{poisson_eta_perf}, the both the blue and red curves have comparable minimum values.
    
    \subsection{Effect of varying the discount factor $\gamma$}
    Recall that in the classical problem formulation, the discount factor $\gamma$ can be interpreted as the weight of future observations. In this section, we investigate whether changing this parameter can overcome the buit-in bias against policies with longer mixing times, without resorting to the $\eta$-optimality criterion. 
    In practice, the discount factor $\gamma$ is used separately in two building blocks of IRL algorithms (including $\GAIL$ and $\MEGAN$). The first instance is in the Bellman updates when solving the RL problem under a given cost function: we refer to this parameter as $\gamma_{\RL}$. The second instance is in the discrimination problem when approximately sampling from $\rho_\pi$: we refer to this parameter as~$\gamma_{\IRL}$. 
    
    Due to the finite nature of the expert demonstrations, the standard approach is to approximate future state distributions by setting $\gamma_{\IRL}$ to $1$ (or equivalently, sampling transitions uniformly). This can be seen as an asymptotic behaviour of the classical problem formulation: as the discount factor approaches $1$, the associated truncated geometric distribution will converge to a uniform one. Reducing this parameter will only accentuate the discussed issues as it entails up-sampling the early stages of the collected demonstrations, which will inevitably favor short term imitation. 
    On the other hand, it is not possible to set $\gamma_{\RL}=1$ (the Bellman operator is no longer guaranteed to admit a fixed point, and state of the art value based RL algorithms become extremely unstable). For this reason, most practitioners set this parameter to a value reasonably close to $1$.
    
    %In order to establish the inability of classical IRL algorithms to match the expert behaviour as well as their generalised counterparts
    In this section, we evaluated the remaining replay buffers obtained using $\GAIL$ as we vary the discount factors values ($\gamma_{RL}\in[0.9, 0.99, 0.999]$). We emphasize that in all the reported empirical evaluations (including previous ones, i.e. Figures \ref{geom_eta_perf} and \ref{poisson_eta_perf}), we fixed $\gamma_{\IRL}$ to $1$.
    
\begin{figure}
    \centering
    \begin{subfigure}[b]{.32\linewidth}
        \centering
        \includegraphics[width=\linewidth]{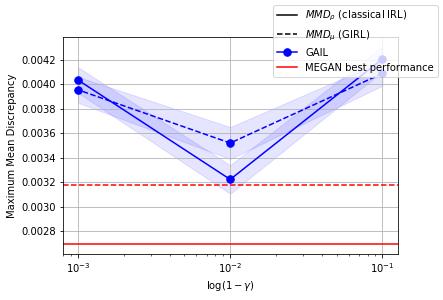}
        \caption{Ant environment}
        \label{gamma_perf_ant}
    \end{subfigure}
    \begin{subfigure}[b]{.32\linewidth}
        \centering
        \includegraphics[width=\linewidth]{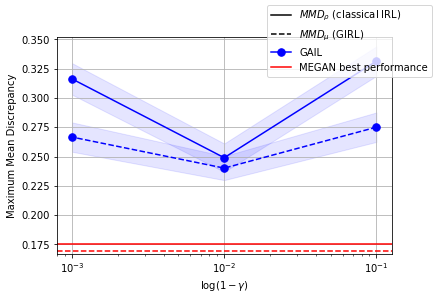}
        \caption{Half-Cheetah environment}
        \label{gamma_perf_half_cheetah}
    \end{subfigure}
    \begin{subfigure}[b]{.32\linewidth}
        \centering
        \includegraphics[width=\linewidth]{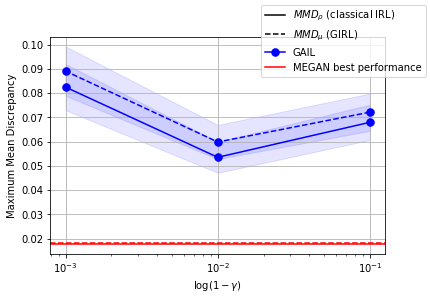}
        \caption{Hopper environment}
        \label{gamma_perf_hopper}
    \end{subfigure}
    \caption{ \textbf{Performances of $\GAIL$ as we vary the discount factor $\gamma$:} Neither increasing nor decreasing the discount factor resulted in improved performances. Agnostic of the used parameter, $\GAIL$ was not able to match the expert behavior as well as $\MEGAN$.}
    \label{gamma_perf}
    \end{figure}

    In Figure \ref{gamma_perf}, we report the average $\MMD_\rho$ divergence of the remaining replay buffer in solid lines, and the average $\MMD_\mu$ divergence in dashed lines. $\GAIL$'s performances as we vary the discount factor are reported in blue and the best performances obtained with $\MEGAN$ are reported in red. Reducing $\gamma_{RL}$ to $0.9$ accentuated the bias against policies with longer mixing times, and on the other hand increasing it to $0.999$ lead to a less reliable RL algorithm. As expected, we observe in Figure \ref{gamma_perf} that both tweaks did not entail performances on par to what we obtained using $\MEGAN$. 
    
\section{Conclusion}
    In this paper, we generalised the classical criterion of optimality in the reinforcement learning literature by putting more weights onto future observations. 
    Using this novel criterion, we reformulated both the regularised RL and the maximum-entropy IRL problems.
    We reviewed existing RL algorithms and discussed their ability to search for $\eta$-optimal policies.
    We also generalised classical IRL solutions. % and empirically verified improved algorithmic stability and enhanced expert matching quality of the obtained imitation policies. 
    The derived algorithm produced stable solutions with enhanced expert matching properties.
    
    In practice, the main difference between MEGAN and GAIL is the discriminator's sampling procedure. This implies that it can easily replace the latter algorithm in all subsequent contributions.
    %Similarly, the proposed criterion can be exploited to generalise existing policy-based RL algorithms. 
    An interesting future direction of research consists in evaluating the margin of improvement that can be gained from this modification.

%%%%%%%%%%%%%%%%%%%%%%%%%%%%%%%%%%%%%%%%%%%%%%%%%%%%%%%%%%%%%%%%%%%%%%%%%%%%%%%%%%%%%%%%%%%%%%%


\bibliography{example_paper}
\bibliographystyle{abbrv}

%%%%%%%%%%%%%%%%%%%%%%%%%%%%%%%%%%%%%%%%%%%%%%%%%%%%%%%%%%%%%%%%%%%%%%%%%%%%%%%
\newpage
\appendix
\onecolumn
\section{Generalisability of policy gradient based RL algorithms}
\label{A:pol_grad}
    Policy gradient based RL algorithms minimise the cumulative discounted cost by directly optimising the policy parameters using gradient descent. We discuss in this section the ability of these algorithms to search for $\eta$-optimal policies. 
    We distinguish two types of policy gradient approaches:
    
    \subsection*{Optimising the global objective directly:}
        Many methods (such as Deterministic Policy Gradient \cite{DPG}, and variants of the REINFORCE algorithm \cite{reinforce_1, reinforce_2}) are rooted in the policy gradient theorem \cite{pol_grad}: 
        \begin{align*}
            & \nabla \mathbb{E}_{p_0, \pi}[\sum_t \gamma^t c(s_t,a_t)]  = \int_{s_0, s} \hspace{-0.4cm} p_0(s_0) \rho_\pi(s|s_0) \mathbb{E}_{a\sim\pi}[Q_\pi^c(s,a) \nabla\log\pi(a|s)] \\
            \text{with}\ & \rho_\pi(s|s_0) = \sum_t \gamma^t\mathbb{P}_\pi(s_t = s | s_0)
        \end{align*}
        where the policy $\pi$ is a function of some parameter $\theta$ and all gradients are implicitly with respect to $\theta$.
        In order to adapt these approaches for the search of \hbox{$\eta$-optimal} policies, the policy updates must take into account the future state distribution derivative. In fact, the gradient of the generalised criterion with respect to the policy induces an additional term as provided in proposition \ref{P:policy_grad_th}:
        \begin{proposition} For any given distribution $\eta$: 
            \begin{align*}
                \nabla \mathds{E}^{\eta}_{p_0,\pi}[\sum_t \gamma^t c(s_t,a_t)] = &
                \underbrace{\int_{s_0, s_+}\hspace{-0.6cm} p_0(s_0) v_{\pi}^c(s_{+}) \nabla P^{\eta}_{\pi}(s_{+}|s_0)}_\textbf{additional term} \\
                & + \underbrace{\int_{s_0, s_+, s}\hspace{-0.85cm} p_0(s_0) P^{\eta}_{\pi}(s_{+}|s_0) \rho_\pi(s|s_{+}) \mathbb{E}_{\pi}[Q_{\pi}^c(s,a) \nabla\log\pi_\theta(a|s)]}_\textbf{modified term} 
            \end{align*}
            \label{P:policy_grad_th}
            where $P_{\pi}^{\eta}(s_{+}|s_0) = \sum_{n=0}^\infty \eta(n) \mathbb{P}_\pi(s_n = s_{+}|s_0)$.
        \end{proposition}
        Notice that the modified term has the same form as the original policy gradient theorem. This is not an issue as it's a matter of adapting the used estimators in practice.
        However, the additional term is not taken into account. Furthermore, in this current form, $\nabla P^{\eta}_{\pi}(s_{+}|s_0)$ is not tractable.
        This implies that current policy gradient approaches that rely on the policy gradient theorem can not search in a reliable way for $\eta$-optimal policies.
    \subsection*{Optimising a local version of the global objective:}
        On the other hand, recent policy gradient algorithms such as Trust Region Policy Optimisation (TRPO) \cite{trpo} and Proximal Policy Optimisation (PPO) \cite{ppo} iteratively search for a new policy $\pi_{n}$ that improves the performances of an old policy $\pi_{o}$ by optimising a local approximation of the right hand term in the following identity \cite{local_improuvement}: 
        \begin{align*}
            \mathcal{L}_0^{\delta_0} (\pi_{n}, c)
            = & \mathcal{L}_0^{\delta_0} (\pi_{o}, c) + \mathbb{E}_{p_0, \pi_{n}}[\sum_t \gamma^t A_{\pi_{o}}^c(s_t,a_t)] \\
            = & \mathcal{L}_0^{\delta_0} (\pi_{o}, c) + \int_{s_0,s}\hspace{-0.4cm} p_0(s_0) \rho_{\pi_{n}}(s|s_0) \int_a \pi_{n}(a|s) A_{\pi_{o}}^c(s,a)
        \end{align*}
        where $\mathcal{L}_0^{\delta_0}$ is the unregulated loss function, and $A_{\pi}^c(s,a)$ is the advantage function:
        \begin{align*}
            A_{\pi}^c(s,a) = Q_{\pi}^c(s,a) - v_{\pi}^c(s) 
        \end{align*}
        In principle, this approach is tractable for the search of $\eta$-optimal policies. In fact, the generalised setting verifies a similar formulation of this identity: 
        \begin{proposition}
            For any given distribution $\eta$:
            \begin{align*}
                    \mathcal{L}_0^{\eta} (\pi_{n}, c) & = \mathcal{L}_0^{\eta} (\pi_{o}, c) + \mathbb{E}_{p_0, \pi_{n}}^\eta[\sum_t \gamma^t A_{\pi_{o}}^c(s_t,a_t)] \\ 
                    & = \;\mathcal{L}_0^{\eta} (\pi_{o}, c) + \int_{s_0,s}\hspace{-0.4cm} p_0(s_0) \rho_{\pi_{n}}(s|s_0)  \mathbb{E}_{\pi_{n}}^{\eta}[A_{\pi_{o}}^c(s_{+},a_{+})|s] \\
                    \text{with } \qquad &\mathbb{E}_{\pi_{n}}^{\eta}[A_{\pi_{o}}^c(s_{+},a_{+})|s] =\hspace{-0.1cm}\int_{s_{+},a_{+}}\hspace{-0.7cm} P_{\pi_{n}}^{\eta}(s_{+}, a_{+}|s) A_{\pi_{o}}^c(s_{+},a_{+})
            \end{align*}
            \label{P:trpo_identity}
        \end{proposition}
        Propositions \ref{P:trpo_identity} lay the ground to adapt proximal policy gradient based RL algorithms to the search of $\eta$-optimal policies. However, we do not investigate this further in this paper as we focus on the Inverse problem.
    
\section{Generalisation of classical IRL algorithms}
\label{A:generalised_algorithms}
    In this section, we discuss particular classical penalisation functions $\Omega$ and $\Psi$ that lead to generalisation of well known $\IRL$ algorithms. In all cases, $\Omega$ is always chosen as the entropy regulariser,  as in state of the art $\RL$ algorithms. 
    
    Let $\mathcal{C}$ be a subset of admissible cost functions, and the penalisation function defined as: 
        \begin{equation*}
            \psi(c) = \imath_{\mathcal{C}}(c) =
            \left\{
            \begin{array}{ll}
                0 & \textit{if } c\in\mathcal{C}  \\
                +\infty & \textit{if } c\notin\mathcal{C}
            \end{array}
            \right.
        \end{equation*}
       Two particular  subsets are studied in details in the following as they  lead to generalisations of classical $\IRL$ algorithms.
    
    \subsection{EMMA: Expectation Matching - Maximum Entropy}
        First, consider the set of linear interpolation of some finite basis set function $\{f_i(s,a), \ i \in \mathcal{I}\}$, i.e.,
        $$\mathcal{C}_{\textit{linear}} = \Big\{ \sum _{i \in \mathcal{I}} w_i f_i, \ \textit{ such that } \|w\|_2 \leq 1 \Big\}.$$
        In the classical, non-generalised IRL problem (with $\eta=\delta_0$), this problem coincides with features expectation matching $\IRL$ algorithm \cite{apprenticeship}, that minimises the $l_2$ expected feature vectors \cite{gail}:
        \begin{align*}
                L(\pi,c) + \Omega(\pi) = \max_{c\in\mathcal{C}_{\textit{linear}}} \mathbb{E}_{\rho_\pi}[c(s,a)]-\mathbb{E}_{\rho_{\bar{\pi}}}[c(s,a)] = \big\| \mathbb{E}_{\rho_\pi}[f]-\mathbb{E}_{\rho_{\bar{\pi}_E}}[f] \big\|_2,
        \end{align*}
        where $f(s,a)=(f_i(s,a))_{i \in \mathcal{I}}$. 
        Generalising this algorithm for any geometric distribution $\eta$ simply consists in substituting the expectation with $\mathds{E}_\pi^\eta$:
        \begin{proposition}
            Under the assumptions of Proposition \ref{PR:equivalent}, and for $\psi=\imath_{\mathcal{C}_{\textit{linear}}}$, it holds that:
            $$
                \RL_\Omega^\eta \circ \IRL_{\psi}^\eta (\bar{\pi}) =  \argmin_\pi -\Omega(\pi) +  \big\|\mathbb{E}_\pi^\eta[f]-\mathbb{E}_{\pi_E}^\eta[f] \big\|_2
            $$            
            \label{P:EMMA}
        \end{proposition}
        \vspace{-.7cm}
        The generalised version of this algorithm (which is actually $\GIRL_{\imath{\mathcal{C}_{\textit{linear}}}, H, \eta}$) that we called $\EMMA_{\eta}$, is derived in the following as a generalisation of expectation matching $\IRL$ algorithm \cite{apprenticeship}. 
        
        The optimal cost function $c^*_\pi$ (given a previously learned policy $\pi$) must satisfy the following equality\footnote{c.f. the proof of proposition \ref{P:EMMA} in Appendix \ref{A:proofs}}: 
        \begin{align}
            \int_{s_0} p_0(s_0) \big[\mu_\pi(s_{+},a_{+}|s_0)-\mu_{\pi_E}(s_{+},a_{+}|s_0)\big]  c^*_\pi(s_{+},a_{+}) = \Big\|\mathbb{E}_{\mu_\pi}[f]-\mathbb{E}_{\mu_{\pi_E}}[f] \Big\|_2
        \end{align}
        As the objective solution is known, we propose to replace the cost optimisation step in the template procedure we provide in Algorithm \ref{GIRL} with the following simple quadratic loss:
        \begin{align}
            \mathcal{L}_\text{linear}(w) = \Big( \mathbb{E}_{\mu_\pi}[wf^T]-\mathbb{E}_{\mu_{\pi_E}}[wf^T] - \|\mathbb{E}_{\mu_\pi}[f]-\mathbb{E}_{\mu_{\pi_E}}[f]\|_2  \Big)^2
        \end{align}
        Given that the set of feasible cost function is convex, we can update the loss using projected gradient updates. We also use an approximation of this loss in practice: 
        \begin{align}
            \Bar{\mathcal{L}}_\text{linear}(w) = \Big( \sum_{S^+,A^+} wf^T(s,a) - \sum_{S^+_E,A^+_E} wf^T(s,a) -  \|\sum_{S^+,A^+} f(s,a) - \sum_{S^+_E,A^+_E} f(s,a)\|_2 \Big)^2
        \end{align}
        The algorithm we propose is then defined as follows: 
        \begin{algorithm}
            \caption{$\EMMA$}
            \begin{algorithmic}[1]
                \STATE {\bfseries Input:} Expert trajectories $\tau_E \sim \pi_E$, initial policy $\pi_{\theta_0}$ and initial cost function $w_0$
                \FOR{$e \in [1, N]$}
                \STATE Sample trajectories $\tau \sim \pi_{\theta_i}$
                \STATE Sample states randomly $(S_t,A_t)\sim \tau$ and $(S^+,A^+) = (S_{t+k}, A_{t+k})$ where $k\sim\eta$  
                \STATE Sample states randomly $(S'_t,A'_t)\sim \tau_E$ and $(S^+_E,A^+_E) = (S'_{t+k}, A'_{t+k})$ where $k\sim\eta$  
                \STATE Update the cost weights $w_i$ to minimise \hbox{$ \Bar{\mathcal{L}}_\text{linear}(w_i)$}
                \STATE Project the cost weights on the feasible set $\mathcal{C}_\text{linear}$
                 \STATE Update $\theta_i$ using soft actor critic to minimise $w_{i+1}f^T$ 
                \ENDFOR
                \STATE {\bfseries Return:} $(\pi_{\theta_N}, D_{w_N})$
            \end{algorithmic}\label{Algo:EMMA}
        \end{algorithm}
    
    \subsection{WIEM: Worst Individual cost - Entropy Maximizer:}
        We now consider convex combination of basis functions:
        $$ \mathcal{C}_{\textit{convex}} = \Big\{ \sum _{i \in \mathcal{I}} w_i f_i, \textit{ with} \sum_{i \in \mathcal{I}} w_i = 1, \text{ and } w_i \geq 0, \forall i 
        \in \mathcal{I}\Big\} $$
        
        In the classical non-generalised IRL setting, this is equivalent to MWAL \cite{MWAL} and LPAL \cite{LPAL} where we minimise the worst-case excess cost among the basis functions \cite{gail}:
        \begin{align*}
                L(\pi,c) + \Omega(\pi) = \max_{c\in\mathcal{C}_{\textit{convex}}} \mathbb{E}_{\rho_\pi}[c(s,a)]-\mathbb{E}_{\rho_{\bar{\pi}}}[c(s,a)] = \max_{i\in\mathcal{I}} \mathbb{E}_{\rho_\pi}[f_i]-\mathbb{E}_{\rho_{\bar{\pi}}}[f_i]
        \end{align*}
        
        This setting is also simply generalised for any geometric $\eta$ by takng the expectation w.r.t.~$\mathds{E}_\pi^\eta$: 
        \begin{proposition}
            Under the assumptions of Proposition \ref{PR:equivalent}, and for $\psi=\imath_{\mathcal{\mathcal{C}_{\textit{convex}}}}$, it holds that:
            \begin{align*}
                     \RL_\Omega^\eta \circ \IRL_{\psi}^\eta (\pi_E) = \argmin_\pi -\Omega(\pi) +
                     \max_i \mathbb{E}_\pi^\eta[f_i]-\mathbb{E}_{\pi_E}^\eta[f_i]
            \end{align*}   
            \label{P:WIEM}          
        \end{proposition}
        \vspace{-.7cm}
        
        We derive $\WIEM_{\eta}$ in the following, which is equivalent to $\GIRL_{\delta_{\mathcal{C}_{\textit{convex}}}, H, \eta}$ and a generalisation of worst-case excess IRL algorithms. 
    
        The optimal cost function $c^*_\pi$ (given a previously learned policy $\pi$) must satisfy the following equality\footnote{c.f. the proof of proposition \ref{P:WIEM} in Appendix \ref{A:proofs}}: 
        \begin{align}
            \int_{s_0} p_0(s_0) \big[\mu_\pi(s_{+},a_{+}|s_0)-\mu_{\pi_E}(s_{+},a_{+}|s_0)\big]  c^*_\pi(s_{+},a_{+}) = \max_i \mathbb{E}_{\mu_\pi}[f_i]-\mathbb{E}_{\mu_{\pi_E}}[f_i] 
        \end{align}
        As the objective solution is known, we propose to replace the cost optimisation step in the template procedure we provide in Algorithm \ref{GIRL} with the following simple quadratic loss:
        \begin{align}
            \mathcal{L}_\text{convex}(w) = \Big( \mathbb{E}_{\mu_\pi}[wf^T]-\mathbb{E}_{\mu_{\pi_E}}[wf^T] - \max_i \mathbb{E}_{\mu_\pi}[f_i]-\mathbb{E}_{\mu_{\pi_E}}[f_i]   \Big)^2
        \end{align}
        Given that the set of feasible cost function is convex, we can update the loss using projected gradient updates. We also use an approximation of this loss in practice: 
        \begin{align}
            \Bar{\mathcal{L}}_\text{convex}(w) = \Big( \sum_{S^+,A^+} wf^T(s,a) - \sum_{S^+_E,A^+_E} wf^T(s,a) - \max_i \big[\sum_{S^+,A^+} f_i(s,a) - \sum_{S^+_E,A^+_E} f_i(s,a)\big] \Big)^2
        \end{align}
        The algorithm we propose is then defined as follows: 
        \begin{algorithm}
            \caption{$\WIEM$}
            \begin{algorithmic}[1]
                \STATE {\bfseries Input:} Expert trajectories $\tau_E \sim \pi_E$, initial policy $\pi_{\theta_0}$ and initial cost function $w_0$
                \FOR{$e \in [1, N]$}
                \STATE Sample trajectories $\tau \sim \pi_{\theta_i}$
                \STATE Sample states randomly $(S_t,A_t)\sim \tau$ and $(S^+,A^+) = (S_{t+k}, A_{t+k})$ where $k\sim\eta$  
                \STATE Sample states randomly $(S'_t,A'_t)\sim \tau_E$ and $(S^+_E,A^+_E) = (S'_{t+k}, A'_{t+k})$ where $k\sim\eta$  
                \STATE Update the cost weights $w_i$ to minimise \hbox{$ \Bar{\mathcal{L}}_\text{convex}(w_i)$}
                \STATE Project the cost weights on the feasible set $\mathcal{C}_\text{convex}$
                 \STATE Update $\theta_i$ using soft actor critic to minimise $w_{i+1}f^T$ 
                \ENDFOR
                \STATE {\bfseries Return:} $(\pi_{\theta_N}, D_{w_N})$
            \end{algorithmic}\label{Algo:WIEM}
        \end{algorithm}
\section{Multi-task setting}
\label{A:multi_task_setting}
    Classically, the multi-task setting is defined by considering a task space $\Theta$ and for each task $\theta\in\Theta$ the associated Markov decision process \hbox{$\mathcal{M}_\theta=\{\mathcal{S}, \mathcal{A}, \mathcal{P}, c_\theta, \gamma, p_0\}$}. Depending on the context, the objective
     is then to either solve the RL or the IRL problems for the set of MDPs $(\mathcal{M}_\theta)_{\theta\in\Theta}$ by averaging the losses with respect to a task distribution $\mathcal{F}$.
        This is equivalent in principle to solving the problem for the MDP \hbox{$\Bar{\mathcal{M}}=\{\mathcal{S}\times\Theta, \Bar{\mathcal{A}}, \Bar{\mathcal{P}}, \Bar{c}, \gamma, \Bar{p}_0\}$} where for any states $(s,s')$, tasks $(\theta,\theta')$ and action $a$, the following equalities hold true:
        \begin{align*}
            &\Bar{\mathcal{A}}(s,\theta)  &= & \; \mathcal{A}(s) & \\
            &\Bar{\mathcal{P}}(s',\theta'|s,\theta,a) &  = & \; \mathcal{P}(s'|s,a)\delta(\theta'=\theta)&\\
            &\Bar{c}(s,\theta,a) & = & \; c_\theta(s,a) & \\
            &\Bar{p}_0(s,\theta) & = & \; p_0(s)\mathcal{F}(\theta) &
        \end{align*}
        We adapt the latter formulation here for the sake of coherence with previous sections.
        
        The difficulty in multi-task settings arises from ray-interference: when the cost function encourages conflicting behaviours for different tasks, the learning objective plateaus \cite{ray_interference}. This stagnates the progress of the policy, which in turn complicates the IRL problem as these plateaus are an opportunity for the discriminator to over-fit the replay-buffer. To alleviate this issue, we propose to \textit{augment the data-set} as proposed in Section \ref{sec:data_augmentation} (by using $\MEGAN$ coupled with the Idle subroutine.
\subsection{Idle : an On-policy data augmentation routine}
    \label{sec:data_augmentation}
    If the state space $\mathcal{S}$ is very large, or even continuous, it becomes quite unlikely to encounter the same state twice in a (finite) trajectory from a given data-set.  In particular, this renders quite difficult the estimation of future state distribution $P_\pi^\eta(.|s)$ (where $s\sim\rho_\pi(.|s_0)$).
    %Intuitively, an expert data-set can be used to solve the IRL problem if it covers wide variety of possible expert trajectories. In other words, sampling transitions from it should generates a reliable approximation of both the occupancy measure $\rho_\pi(.|s_0)$ (where $s_0\sim p_0$) and the future state distribution $P_\pi^\eta(.|s)$ (where $s\sim\rho_\pi(.|s_0)$). Particularly in continuous settings, and as it becomes extremely rare to encounter the same states multiple times, the latter condition is harder to achieve (or at least it would require exponentially more trajectories). 
    
    To circumvent this issue, we propose to use the following on-policy data augmentation scheme, modeled as a game between a discriminator \hbox{$D: (\mathcal{S}\times\mathcal{A}\times\mathcal{S})\to[0,1]$} and a generator~\hbox{$G:\mathcal{S}\rightarrow\Delta(\mathcal{S}\times\mathcal{A})$}. The objective of the generator is to produce future states similar to the gathered samples while the discriminator $D$ aims to identify true samples from generated ones, with  the following score function:
    % \begin{align*}
    %     V(D,G) = \mathbb{E}\big[ & \log(D(s_+,a_+|s)) + \\
    %     & \log(1-D(s_g,a_g|s)) \big] 
    % \label{idle_game_on_policy}
    % \end{align*}
       \begin{equation*}
        V(D,G) = \mathbb{E}\big[  \log(D(s_+,a_+|s)) + 
        \log(1-D(s_g,a_g|s)) \big] 
   % \label{idle_game_on_policy} %VP:aucun interet si align*
    \end{equation*}
    where the expectation is taken w.r.t the future state distribution $P_\pi^\eta(.|s)$ for $(s_+,a_+)$, the generator distribution $G(.|s)$ for $(s_g,a_g)$, and the marginal  over the initial state distribution of the occupancy measure $\rho_\pi(.|s_0)$ for $s$. Solving this game approximates $P_{\pi}^{\eta}$: 
    
    \begin{proposition}
        $(\Tilde{D},\Tilde{G})= (\frac{1}{2},P_{\pi}^{\eta})$ is a Nash-equilibrium of the following zero-sum game:
        \begin{align}
            D^*: \  \min_D V(D,G) \; \ \text{ and }  \;
            G^*: \ \max_G V(D,G)
        \end{align}
        \label{P:Idle}
    \end{proposition}
    \vspace{-.7cm}
    From this proposition, we derive in Algorithm \ref{A:Idle_On_policy}, a method to approximate future state distributions that is used as a subroutine in $\GIRL$. This approach is theoretically feasible given an on-policy data set (such as the expert's trajectories). In practice, we noticed that approximating $P_\pi^\eta$ using (Idle) produces reliable generators when the variance of $\eta$ is relatively small, so that future state samples fall within a reduced range with a high probability. Inversely, if $\eta$ has a high variance, samples from $P_\pi^\eta$ would be along extended horizon and the (Idle) discriminator easily picks up on the parts being learned and halts the generator’s improvement. This either leads to vanishing gradient updates or a mode collapse.
    
    \begin{algorithm}
            \caption{Idle (an on-policy future state generator)}\label{A:Idle_On_policy}
            \begin{algorithmic}[1]
                \STATE {\bfseries Input:} On-policy trajectories $\tau$, initial discriminator $D_{\phi_0}$ and initial generator $G_{\nu_0}$
                \FOR{$e \in [1, N]$}
                \STATE Sample states randomly $(S_t,A_t)\sim \tau$
                \STATE Sample $(S^+,A^+) = (S_{t+k}, A_{t+k})$ where $k\sim\eta$  
                \STATE Sample $(S^+_G,A^+_G)\sim G_{\nu_i}(S)$
                \STATE Update the discriminator parameter $\phi_i$ to minimise: 
                \begin{align*}
                    \sum_{\substack{S_t,S^+,A^+}} \log(D_{\phi_i}(s_+,a_+|s)) + \sum_{\substack{S_t,S^+_G,A^+_G}} \log(1-D_{\phi_i}(s_+,a_+|s))
                \end{align*}
                 \STATE Update the generator parameter $\nu_i$ to minimise: 
                  $\quad\quad \sum_{\substack{S_t}} \log(D(G(s)|s))$
                \ENDFOR
                \STATE {\bfseries Return:} $(D_{\phi_N}, G_{\nu_N})$
            \end{algorithmic}
    \end{algorithm}

\section{Experiments for the Multi-task setting and the idle procedure}
\label{A:experiments_multi-task}
    \subsection{Fetch-Reach environment}
        We consider in this section the \hbox{\textit{FetchReach}}\footnote{To the extent of our knowledge, this is the first reported performances of IRL algorithms on a fully continuous environment} task from the \textit{MuJoCo} based environments \cite{mujoco_gym}.  
        To evaluate the generalisability of the learned policies, we only generate expert trajectories for a subset of possible tasks (only target positions that are $5$-$10$ cm away from the initial gripper's position\footnote{the maximum range of the arm is about $25$ cm} to be precise). We evaluate the learned policies in the \textit{learned setting} (same horizon and same tasks) and in a \textit{generalisability setting} (twice the training horizon and the full range of tasks). As in the simple task setting, we asses performances in terms of normalised cumulative costs.
        
        \begin{figure}
       \centering
            \begin{subfigure}{.49\linewidth}
                \includegraphics[width=\linewidth]{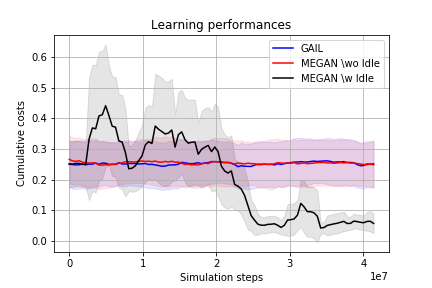}
                \caption{Normalised cumulative costs over training}
                \label{multi_task_learning}
            \end{subfigure}
            \begin{subfigure}{.46\linewidth}
                \includegraphics[width=.9\linewidth]{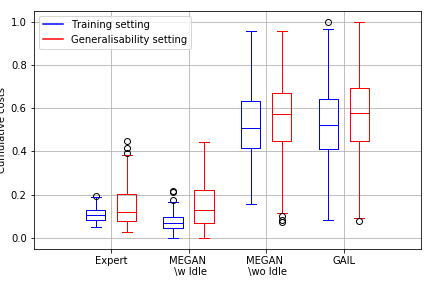}
                \caption{Normalised cumulative costs (best performance)}
                \label{multi_task}
            \end{subfigure}
            \caption{Performances in the Fetch Reach setting}
            \label{multi-task}
        \end{figure} 
        
        In Figure \ref{multi_task_learning}, we compare the performances of MEGAN (with and without the data augmentation) and GAIL over the training. We observe that both GAIL and MEGAN (without Idle) struggle in solving the problem. However, using the Idle generator reduces the undesirable effect of ray interference and stabilises the training.
        Performance wise, the learned policy using MEGAN outperforms the expert demonstrations in the training tasks while at the same time providing comparable performances in the remaining set of tasks (as provided in Figure \ref{multi_task}).
        
        \begin{wrapfigure}[15]{r}{0.4\textwidth}
            \includegraphics[width=\linewidth]{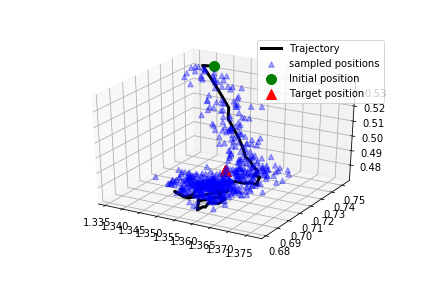}
            \caption{Idly generated samples from expert trajectory}
            \label{idle_samples}
        \end{wrapfigure}
        It is arguable that the success of MEGAN in the multi-task setting is explained with the Idle procedure. A similar approach on GAIL might be appealing. However, this is not feasible in practice.
        It is true that the reasoning provided in Section \ref{sec:data_augmentation} can be developed for any distribution $\eta$, this entails that we can use the same approach to learn a generator that mimics $\rho_\pi(.|s_0)$. Unfortunately, this is not feasible in practice (due to the high variance of $\rho_\pi$, as explained in Section \ref{sec:data_augmentation}). The issue at play here is that the latter distribution ($\rho_\pi)$ covers all the observations of the trajectory. Indeed, from our early empirical attempts, we noticed that learning such distribution is unstable: we either over-fit a sub-set of the trajectory (notably the stationary distribution, which is hurtful for our purpose) or we do not learn the distribution at all. 
        On the other hand, learning $P_\pi^\eta$ conditioned on some intermediate state is a lot easier when we choose $\eta$ such that it only covers the near future transitions (for example when $\eta=\Geom(0.7)$, the average prediction of $P_\pi^\eta(|s_t)$ is $3$ steps in the future). 
        
        In Figure \ref{idle_samples}, we evaluate the learned approximation of $P_{\pi_E}^{\Geom(0.7)}$ on a sample expert trajectory. We plot the evolution of  the (true/generated) gripper position in 3D overtime.  For each state encountered on the trajectory, the learned generator outputs $10$ samples (in blue). Clearly, the future state generator is reliable; this is successful because the distribution $\eta$ is a short term prediction: the learned generator maps current states to the possible ones in the next few steps.

    \subsection{Multi-task 2-D navigation}
        \begin{wrapfigure}[15]{r}{0.3\textwidth}
            %\centering
            \includegraphics[width=0.3\textwidth]{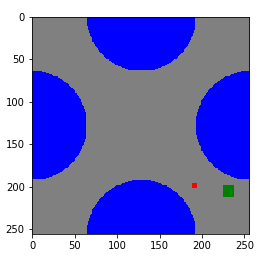}
            \caption{2-D navigation environment}
        \label{fig:2d_nav}
        \end{wrapfigure}
\begin{figure}
            \centering
                \begin{subfigure}[b]{.3\linewidth}
                \centering
                    \includegraphics[width=\linewidth]{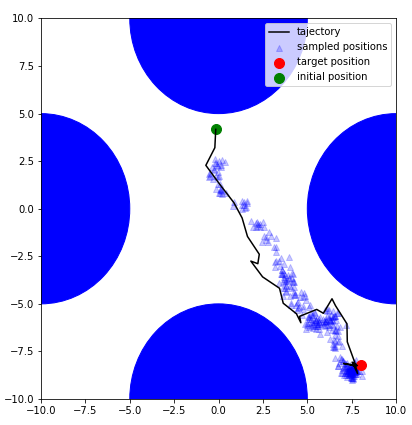}
                \end{subfigure}
                \begin{subfigure}[b]{.3\linewidth}
                \centering
                    \includegraphics[width=\linewidth]{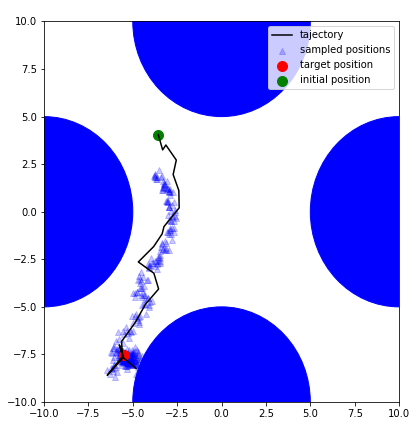}
                \end{subfigure}
                \begin{subfigure}[b]{.3\linewidth}
                \centering
                    \includegraphics[width=\linewidth]{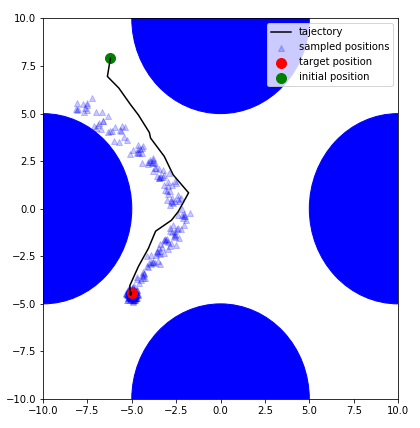}
                \end{subfigure}
            \caption{$P_\pi^\eta$ learned generator }
            \label{P_eta_pplus}
            \end{figure}
\begin{figure}
            \centering
                \begin{subfigure}[b]{.3\linewidth}
                \centering
                    \includegraphics[width=\linewidth]{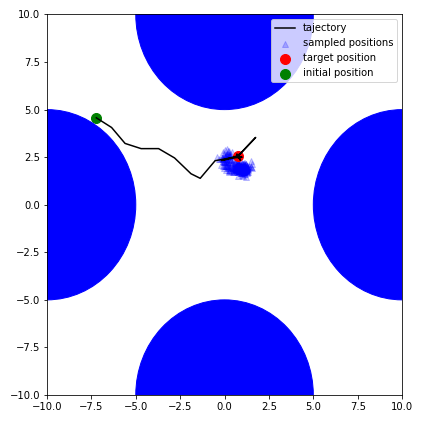}
                \end{subfigure}
                \begin{subfigure}[b]{.3\linewidth}
                \centering
                    \includegraphics[width=\linewidth]{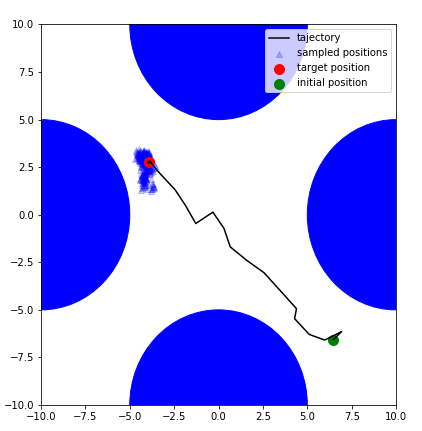}
                \end{subfigure}
                \begin{subfigure}[b]{.3\linewidth}
                \centering
                    \includegraphics[width=\linewidth]{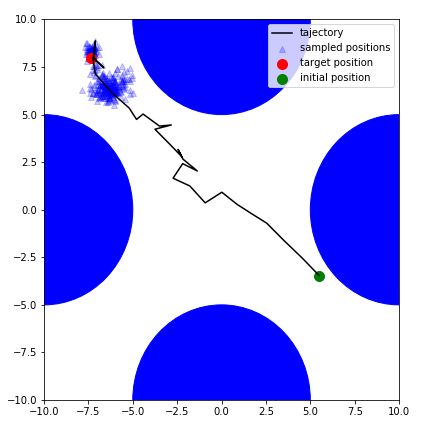}
                \end{subfigure}
            \caption{$\rho_\pi$ learned generator }
            \label{rho_pplus}
            \end{figure}

        In this section we report the performances on a custom-made multi-task navigation environment. The goal is to navigate from an initial position to a target position while avoiding four lakes. The state space is constructed by concatenating the coordinates of the agent, the coordinates of the target as well as the distance from the centre of each of the four lakes. The action space is the norm 2 ball $\{x\in\mathbb{R}^2 \textbf{ s.t.\ } \|x\|_2\leq 1 \}$. The transition kernel is a Dirac mass at the sum of the previous position and the action vector. If the sum is within one of the lakes or outside the grid, then the new position is the projection of the previous position on the border according to the action direction. 
        In Figure \ref{fig:2d_nav}, we render the environment to provide an idea about the task at hand: the lakes are painted in blue, the agent is the red square, the target is the green square, and the grey pixels are the possible positions. These positions are the subset of $[-10,10]^2$ that excludes the points within the lakes. The goal is to navigate around the lakes in order to reach the target position. 
        \paragraph{Learning the Idle generator}
            In this section we analyse the ability to learn $P_\pi^\eta$ and $\rho_\pi$ using Algorithm \ref{A:Idle_On_policy}. As discussed in Section \ref{sec:data_augmentation}, when the target distribution is a short term prediction of future states, the obtained generator is reliable. We consider in what follows $\eta=\Geom(0.7)$ to satisfy this condition. 
            We use the same hyper-parameters to learn the generator in both cases ($P_\pi^\eta$ and $\rho_\pi$). We use 3-layers deep, 64-neurons wide neural networks for the generator and the discriminator, a batch size of $256$ and we iterate the algorithm for $10000$ steps.
            In~Figures \ref{P_eta_pplus} and \ref{rho_pplus}, we plot the expert trajectories with black lines, the initial position with green dots, the target position with red dots, and the sampled states with blue triangles. On one hand, we observe in Figure \ref{P_eta_pplus} that the $P_\pi^\eta$ learned generator provides reliable samples (in the sense that they follow the trails of expert trajectories). On the other hand, in Figure \ref{rho_pplus}, we observe that the learned $\rho_\pi$ generator over-fits the stationary distribution and only samples states around the target position. The mode collapse is essentially explained by the fact that most of the samples from the $\rho_\pi$ distribution are indeed around the target position. 
            
        \paragraph{Learned cost function}
            
\begin{figure}
            \centering
                \begin{subfigure}[b]{.32\linewidth}
                \centering
                    \includegraphics[width=\linewidth]{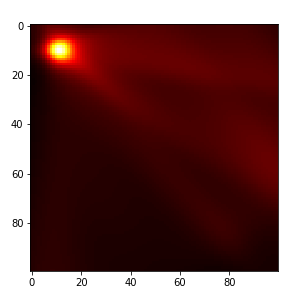}
                    %\caption{ Top left }
                \end{subfigure}
                \begin{subfigure}[b]{.3\linewidth}
                \centering
                    \includegraphics[width=\linewidth]{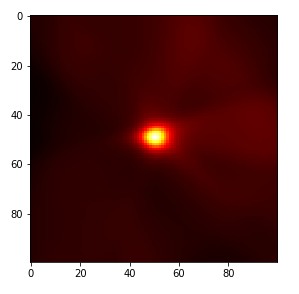}
                    %\caption{ Center }
                \end{subfigure}
                \begin{subfigure}[b]{.3\linewidth}
                \centering
                    \includegraphics[width=\linewidth]{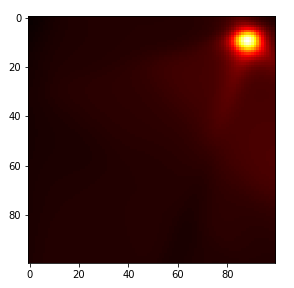}
                    %\caption{ Top ight }
                \end{subfigure}
                \begin{subfigure}[b]{.3\linewidth}
                \centering
                    \includegraphics[width=\linewidth]{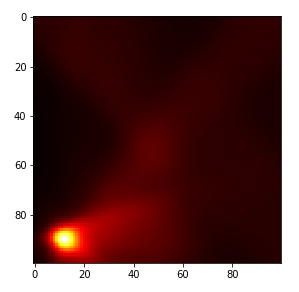}
                    %\caption{ Bottom left }
                \end{subfigure}
                \begin{subfigure}[b]{.3\linewidth}
                \centering
                    \includegraphics[width=\linewidth]{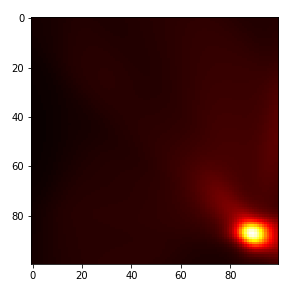}
                    %\caption{ Bottom right }
                \end{subfigure}
            \caption{ GAIL learned cost heat-map as a function of the target position  }
            \label{gail_cost}
            \end{figure}

\begin{figure}
            \centering
                \begin{subfigure}[b]{.3\linewidth}
                \centering
                    \includegraphics[width=\linewidth]{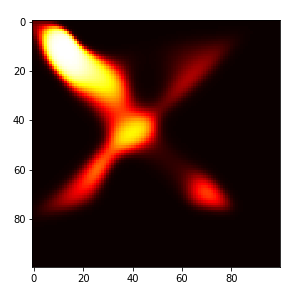}
                    %\caption{ Top left }
                \end{subfigure}
                \begin{subfigure}[b]{.3\linewidth}
                \centering
                    \includegraphics[width=\linewidth]{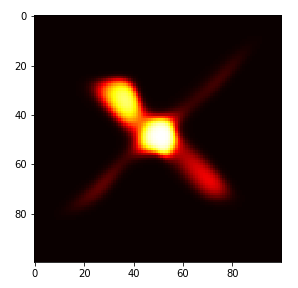}
                    %\caption{ Center }
                \end{subfigure}
                \begin{subfigure}[b]{.29\linewidth}
                \centering
                    \includegraphics[width=\linewidth]{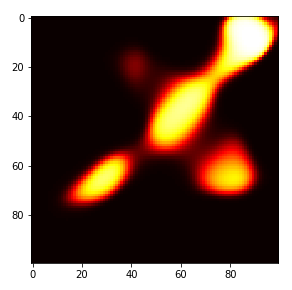}
                    %\caption{ Top ight }
                \end{subfigure}
                \begin{subfigure}[b]{.31\linewidth}
                \centering
                    \includegraphics[width=\linewidth]{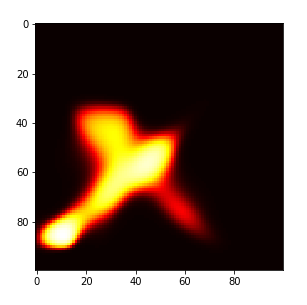}
                    %\caption{ Bottom left }
                \end{subfigure}
                \begin{subfigure}[b]{.3\linewidth}
                \centering
                    \includegraphics[width=\linewidth]{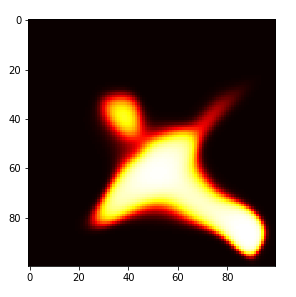}
                    %\caption{ Bottom right }
                \end{subfigure}
            \caption{ MEGAN learned cost heat-map as a function of the target position  }
            \label{megan_cost}
            \end{figure}

            As discussed in Section \ref{sec:experiments}, we only obtain expert-like performances when using the Idle generator. However, given that the considered state-space in this section is a 2-D plan, we can visualise the learned (state only) cost function with a heat-map. We consider five particular tasks that coincide with reaching the top-left, center, top-right, bottom-left and bottom-right of the map. Figures \ref{gail_cost} and \ref{megan_cost} coincide with such heat-maps, with darker shades for higher costs and brighter colours for lower ones. In Figure \ref{gail_cost}, we observe that the GAIL learned costs are particularly low in the vicinity of the target position while they are evenly spreaded elsewhere. This entails from the discriminator over-fitting the replay-buffer as most of the observations are drawn from the stationary distribution. On the other hand, in Figure \ref{megan_cost}, we observe that the MEGAN learned costs are high outside of the paths that lead to the target, and decrease exponentially as we get closer to the goal position. 
            
            %%%%%%%%%

\section{Additional experimental details}
\label{A:experiments}
\subsection{Maximum Mean Discrepancy evaluation}
\label{A:MMD}
    Formally, given a  reproducing kernel Hilbert space (RKHS) of real-valued functions $\mathcal{H}$, the MMD between two distributions $P$ and $Q$ is defined as: $\MMD_{\mathcal{H}}(P, Q) = \sup_{f\in\mathcal{H}} \mathbb{E}_{X\sim P}[f(X)] - \mathbb{E}_{Y\sim Q}[f(Y)]$. Recall that the reproducing property of RKHS, implies that there is a one to one correspondence between positive definite kernels~$k$ and RKHSs $\mathcal{H}$ such that every function $f\in\mathcal{H}$ verifies $f(x)=\langle f,k(.,x)\rangle_{\mathcal{H}}$ (where $\langle\,,\rangle_{\mathcal{H}}$ denotes the RKHS inner product). We propose to evaluate the MMD using a kernel two-sample test with the following unbiased estimator~\cite{MMD}:
    \begin{align*}
        \MMD_{\mathcal{H}}^2(P, Q) = \frac{1}{N(N-1)} \sum_{i\neq j} k(x_i, x_j) +  \frac{1}{N(N-1)} \sum_{i\neq j} k(y_i, y_j) - \frac{1}{N^2} \sum_{i, j} k(x_i, y_j)
    \end{align*}
    where $(x_i)_{i=0}^N$ are sampled according to $P$ and $(y_i)_{i=0}^N$ are sampled according to $Q$.
    In the experimental analysis, we only consider the RKHS associated with the radial basis function $k(x,y) = \exp(\|x-y\|^2/d)$ (where $d$ is the dimension of the variables $x$ and $y$). 

\subsection{Hyper-parameters}
\label{A:Hyper_parameters}
    In this section we provide a detailed description of the used implementation as well as the selected hyper-parameters. 
    
    Expert demonstrations of length \textsc{Max-Length} are stored in a demonstrator replay-buffer.
    We use two additional replay-buffers (one for the policy and one for the expert), with a maximum capacity of $10^6$ transitions that are initially empty. 
    In each cycle, \textsc{N} trajectories from the demonstrator replay-buffer are sampled and added to the expert replay-buffer. The policy generates then \textsc{N} trajectories, that are stored in the policy replay-buffer. In the multi-task setting, the tasks of these trajectories are the same ones in the expert's samples. The policy is updated each cycle using SAC for \textsc{SAC-Epoch} epochs with a bath size of \textsc{SAC-Batch}. Every \textsc{D-Update-Rate} cycles, the discriminator is updated for \textsc{D-Epoch} epochs with a bath size of \textsc{D-Batch}. The algorithm runs until the policy generates \textsc{Max-Transitions} transitions in total.
    
    The policy, as well as the underlying value functions, are approximated using an \textsc{N-Layer-P} deep, \textsc{Hidden-P} wide neural networks. The discriminator is approximated using an \textsc{N-Layer-D} deep, \textsc{Hidden-D} wide neural network.
    
    \begin{center}
    \begin{tabular}{|c||c|c|c|c|c|}
        \hline
        \textsc{Env-ID}  & \textit{Hopper} & \textit{Half-Cheetah} & \hspace{0.1cm} \textit{Ant} \hspace{0.1cm} & \textit{FetchReach} & \textit{2-D Maze} \\
        \hline
        \textsc{N} & 90 & 90 & 90 & 90 & 90\\
        \hline
        \textsc{Max-Length} & 500 & 500 & 500 & 100 & 50 \\
        \hline
        \textsc{SAC-Epoch}& 500 & 500 & 500 & 300 & 150 \\
        \hline
        \textsc{SAC-Batch}& 256 & 256 & 256 & 1024 & 128 \\
        \hline
        \textsc{D-Update-Rate}& 1 & 1  & 1 & 5 & 1 \\
        \hline
        \textsc{D-Epoch}& 50 & 50 & 50 & 500 & 300\\
        \hline
        \textsc{D-Batch}& 512 & 512 & 512 & 512 & 128\\
        \hline
        \textsc{N-Layer-P}& 3 & 3 & 3 & 4 & 4\\
        \hline
        \textsc{Hidden-P}& 64 & 64 & 64 & 64 & 64 \\
        \hline
        \textsc{N-Layer-D}& 1 & 1 & 1 & 3 & 3\\
        \hline
        \textsc{Hidden-D}& 16 & 32 & 32 & 16 & 16 \\
        \hline
        \textsc{Max-Transitions} & $10^7$ & $10^7$ & $10^7$ & $10^6$ &  $5\times10^5$ \\
        \hline
    \end{tabular}
    \end{center}
\section{Proof of technical results}
\label{A:proofs}
    We provide in this section proofs for all stated technical results. To find a particular one, please refer to the following:
    \begin{description}
        \item[Section \ref{A:intermediate_results}:] Useful intermediate results as well as their proof.
        \item[Section \ref{A:GRL}:] Proofs for the theoretical claims stated in Section \ref{sec:GRL} and Appendix \ref{A:pol_grad}.
        \item[Section \ref{A:GIRL}:] Proofs for the theoretical claims stated in Section \ref{sec:GIRL}. 
        \item[Section \ref{A:Idle_data_augmentation}:] Proof for the theoretical claims  stated in Section \ref{sec:data_augmentation}.
        \item[Section \ref{A:particular_settings}:] Proof for the theoretical claims  stated in Section \ref{sec:settings}.
    \end{description}
    
    \subsection{Useful intermediate results}
    \label{A:intermediate_results}
    For the sake of conciseness, we start by providing important intermediate results that will be used in the proofs of~propositions \ref{P:trpo_identity}, \ref{PR:equivalent}, and \ref{P:convex_entropy}. The first one (Proposition \ref{tower_rule}) transforms $\eta$-weighted $\gamma$ discounted functional averaged over $\pi$-generated trajectory into expectations with respect to $\rho_\pi$ and $P_\pi^\eta$ or, equivalently into expectations with respect to $\mu_\pi$. The second one (Proposition \ref{policy_occupancy}) guarantees a one on one mapping between occupancy measures and policies.
    
    \begin{proposition}
        For any distribution $\eta$, and for any mapping $f:\mathcal{S}\times\mathcal{A}\rightarrow\mathbb{R}$, the following identity holds: 
        \begin{align}
            \mathbb{E}_{p_0,\pi}^\eta [\sum_t \gamma^t f(s_t,a_t)] & = \int_{s_0,s,a,s_{+},a_{+}}\hspace{-1cm} p_0(s_0)\rho_\pi(s,a|s_0)P_\pi^\eta(s_{+},a_{+}|s,a) f(s_{+},a_{+}) \\
            & = \int_{s_0,s_{+},a_{+}}\hspace{-.5cm} p_0(s_0)\mu_\pi(s_{+},a_{+}|s_0) f(s_{+},a_{+})
        \end{align}
        \label{tower_rule}
    \end{proposition}
    
    \begin{proposition}
        Let $(\phi_t)_{t=0}^\infty$ be a strictly positive real valued convergent series (i.e. $\sum_t \phi_t<\infty$ and $\phi_t>0$), and let $\Phi_\pi(s,a|s_0)$ be the $\phi$-weighted occupancy measure associated to the policy $\pi$: 
        \begin{align*}
            \Phi_\pi(s,a|s_0) := \sum_t \phi_t \mathbb{P}_\pi(s_t=s,a_t=a|s_0)      
        \end{align*}
        Then for a given $\phi$-weighted occupancy measure $\Phi\in\{\Phi_\pi | \pi:\mathcal{S}\rightarrow\Delta(\mathcal{A})\}$:
        \begin{description}
            \item[1-] $\Phi$ is the $\phi$-weighted occupancy measure of $\pi_\Phi:=\frac{\Phi(s,a|s_0)}{\int_{a'}\Phi(s,a'|s_0)}$
            \item[2-] $\pi_\Phi$ is the only policy whose $\phi$-weighted occupancy measure is $\Phi$
        \end{description}
        \label{policy_occupancy}
    \end{proposition}
    
        \subsubsection*{proof of proposition \ref{tower_rule}:} 
            The proof of the first equality relies on some algebraic manipulations and the law of total expectation. 
            \begin{align}
                & \mathbb{E}_{p_0,\pi}^\eta [\sum_t \gamma^t f(s_t,a_t)] := \int_{s_0,s,a}\hspace{-0.5cm} p_0(s_0) P_\pi^\eta(s,a|s_0) \mathbb{E}_{\pi, \delta_{(s,a)}}[\sum_t \gamma^t f(s_{t},a_{t})] \\
                & \qquad =\int_{s_0,s,a}\hspace{-0.5cm} p_0(s_0)\sum_k \eta(k) \mathbb{P}_\pi(s_k=s,a_k=a|s_0) \mathbb{E}_{\pi}[\sum_t \gamma^t f(s_{t+k},a_{t+k})|s_k=s,a_k=a] \\
                & \qquad = \int_{s_0,s_k,a_k}\hspace{-0.8cm} p_0(s_0)\sum_{k,t} \gamma^t \eta(k) \mathbb{E}_{\pi}\Big[ \mathbb{E}_{\pi}\big[f(s_{t+k},a_{t+k})|s_k,a_k\big]\Big|s_0\Big]
            \end{align}
            where $\mathbb{E}_{\pi, \delta_{(s,a)}}$ designate the expectation over trajectories initialised at the state action couple $(s,a)$. Using the law of total expectation we can assert that: 
            \begin{align}
                \mathbb{E}_{\pi}\Big[ \mathbb{E}_{\pi}\big[f(s_{t+k},a_{t+k})|s_k,a_k\big]\Big|s_0\Big] & = \mathbb{E}_\pi[f(s_{t+k},a_{t+k})|s_0] \\
                & = \mathbb{E}_{\pi}\Big[ \mathbb{E}_{\pi}\big[f(s_{t+k},a_{t+k})|s_t,a_t\big]\Big|s_0\Big]
            \end{align}
            From this relationship, it follows that: 
            \begin{align}
                &\mathbb{E}_{p_0,\pi}^\eta [\sum_t \gamma^t f(s_t,a_t)] = \int_{s_0,s_t,a_t}\hspace{-0.8cm} p_0(s_0)\sum_{k,t} \gamma^t \eta(k) \mathbb{E}_{\pi}\Big[ \mathbb{E}_{\pi}\big[f(s_{t+k},a_{t+k})|s_t,a_t\big]\Big|s_0\Big] \\
                & \qquad = \int_{s_0,s,a}\hspace{-0.5cm} p_0(s_0)\sum_{t} \gamma^t \mathbb{P}_\pi(s_t=s,a_t=a|s_0) \mathbb{E}_{\pi, \delta_{(s,a)}}[\sum_k \eta(k) f(s_{k},a_{k})] \\
                & \qquad = \int_{s_0,s,a}\hspace{-0.5cm} p_0(s_0) \rho_\pi(s,a|s_0) \mathbb{E}_{\pi, \delta_{(s,a)}}[\sum_k \eta(k) f(s_{k},a_{k})] \\
                & \qquad = \int_{s_0,s,a}\hspace{-0.5cm} p_0(s_0) \rho_\pi(s,a|s_0) \sum_k \eta(k) \int_{s_+,a_+}\hspace{-0.5cm}\mathbb{P}_\pi(s_{t+k}=s_{+},a_{t+k}=a_{+}|s_t=s, a_t=a) f(s_+,a_+) \\
                & \qquad = \int_{s_0,s,a,s_{+},a_{+}}\hspace{-1cm} p_0(s_0)\rho_\pi(s,a|s_0)P_\pi^\eta(s_{+},a_{+}|s,a) f(s_{+},a_{+})
            \end{align}
            This concludes the proof of the first equality in Proposition \ref{tower_rule}.
            
            The proof of the second equality relies on the Markov property of the environment and some algebraic manipulations. 
            \begin{align}
                &\mathbb{E}_{p_0,\pi}^\eta [\sum_t \gamma^t f(s_t,a_t)] = \int_{s_0,s,a,s_{+},a_{+}}\hspace{-1cm} p_0(s_0) \rho_\pi(s,a|s_0)P_\pi^\eta(s_{+},a_{+}|s,a) f(s_{+},a_{+}) \\
                & \qquad= \int_{s_0,s,a,s_{+},a_{+}}\hspace{-1cm} p_0(s_0) \sum_{t} \gamma^t \mathbb{P}_\pi(s_t=s, a_t=a|s_0) \sum_k \eta(k) \mathbb{P}_\pi(s_{t+k}=s_+, a_{t+k}=a|s_t=s, a_t=a)f(s_{+},a_{+}) \\
                & \qquad= \int_{s_0,s_{+},a_{+}}\hspace{-.5cm} p_0(s_0) \sum_{t,k} \gamma^t \eta(k) \int_{s,a} \Big( \mathbb{P}_\pi(s_t=s, a_t=a|s_0)  \mathbb{P}_\pi(s_{t+k}=s_+, a_{t+k}=a|s_t=s, a_t=a) \Big) f(s_{+},a_{+}) \\
                & \qquad = \int_{s_0,s_{+},a_{+}}\hspace{-.5cm} p_0(s_0) \sum_{t,k} \gamma^t \eta(k) \mathbb{P}_\pi(s_{t+k}=s_+, a_{t+k}=a|s_0) f(s_{+},a_{+}) \\
                & \qquad = \int_{s_0,s_{+},a_{+}} \hspace{-.5cm} p_0(s_0) \mu_\pi(s_{+},a_{+}|s_0) f(s_{+},a_{+})
            \end{align}
            This concludes the proof of Proposition \ref{tower_rule}.
            
        \subsubsection*{proof of proposition \ref{policy_occupancy}:} 
            For the first assertion of the proposition, recall that:
            \begin{align}
                \Phi_\pi(s,a|s_0) & := \sum_t \phi_t \mathbb{P}_\pi(s_t=s,a_t=a|s_0) \\
                & = \sum_t \phi_t \mathbb{P}_\pi(s_t=s|s_0) \pi(a|s) :=\Phi_\pi(s|s_0) \pi(a|s) 
            \end{align}
            This implies that: 
            \begin{align}
                \frac{\Phi(s,a|s_0)}{\int_{a'}\Phi(s,a'|s_0)} &= \frac{\pi(a|s) \Phi_\pi(s|s_0)}{\int_{a'}\pi(a'|s) \Phi_\pi(s|s_0)} = \pi(a|s)
            \end{align}
            For the second assertion of the proposition, consider two policies $\pi_1$ and $\pi_2$ such that $\Phi_{\pi_1}=\Phi_{\pi_2}$. Notice that: 
            \begin{align}
                \forall s,s_0\in\mathcal{S}& & \Phi_{\pi_1}(s|s_0) & := \sum_t \phi_t \mathbb{P}_{\pi_1}(s_t=s|s_0)  = \int_a \Phi_{\pi_1}(s,a|s_0) \\
                & & & = \int_a \Phi_{\pi_2}(s,a|s_0) = \Phi_{\pi_2}(s|s_0)
            \end{align}
            This can further yield: 
            \begin{align}
                &\forall s\in\mathcal{S}, a\in\mathcal{A} & \Phi_{\pi_1}(s,a|s_0) =  \Phi_{\pi_2}(s,a|s_0) \\
                \Rightarrow&\forall s\in\mathcal{S}, a\in\mathcal{A} & \Phi_{\pi_1}(s|s_0)\pi_1(a|s) =  \Phi_{\pi_2}(s|s_0)\pi_2(a|s) \\
                \Rightarrow & \forall s\in\mathcal{S}, a\in\mathcal{A} & \pi_1(a|s) =  \pi_2(a|s)
            \end{align}
            This concludes the proof of Proposition \ref{policy_occupancy}.
    
    \subsection{Generalised Reinforcement Learning}
    \label{A:GRL}
        In this section we address the claims stated in section \ref{sec:GRL} as well as those stated in Appendix \ref{A:pol_grad}. Proposition \ref{PR:opt_reg_policy} is recalled for the sake of comprehensiveness, a detailed proof is provided in \cite{regMDP}. 
        \subsubsection*{Proof of Corollary \ref{C:gen_opt_reg_policy}:}
            The proof relies on the fact that optimal regularised policies are associated to the minimum regularised value function (as stated in Proposition \ref{PR:opt_reg_policy}). By construction $\pi^*_\Omega$ would minimise the Q-function for all possible state-actions. Recall that for any given policy $\pi$, the following identity hold \cite{regMDP}: 
            \begin{align}
                Q_{\pi,\Omega}^c(s,a) & := Q_\pi^c(s,a) - \Omega(\pi) = c(s,a) + \mathbb{E}_{s'|s,a}[v_{\pi,\Omega}^c(s')]
                \label{regularised_Q_def}
            \end{align}
            From Proposition \ref{PR:opt_reg_policy} and Equation \ref{regularised_Q_def} we can derive the following implications for any policy $\pi$: 
            \begin{align}
                & \forall s\in\mathcal{S} & v_{\pi_\Omega^*,\Omega}^c(s) \leq v^c_{\pi, \Omega}(s) \\
                \Rightarrow & \forall s\in\mathcal{S}, a\in\mathcal{A} & \mathbb{E}_{s'|s,a}[v_{\pi_\Omega^*,\Omega}^c(s')] \leq \mathbb{E}_{s'|s,a}[v_{\pi,\Omega}^c(s')] \\
                \Rightarrow & \forall s\in\mathcal{S}, a\in\mathcal{A} & Q_{\pi_\Omega^*}^c(s,a) - \Omega(\pi_\Omega^*) \leq Q_\pi^c(s,a) - \Omega(\pi) \\
                \Rightarrow & \forall \eta\in\Delta(\mathbb{N}) & \mathds{E}^{\eta}_{p_0,\pi_\Omega^*}[Q^c_{\pi_\Omega^*}] - \Omega(\pi_\Omega^*) \leq \mathds{E}^{\eta}_{p_0,\pi}[Q^c_{\pi}] - \Omega(\pi) \\
                \Rightarrow & \forall \eta\in\Delta(\mathbb{N}) & \pi_\Omega^* \in \argmin_\pi \mathds{E}^{\eta}_{p_0,\pi}[Q^c_{\pi}] - \Omega(\pi) 
            \end{align}
            This concludes the proof of Corollary \ref{C:gen_opt_reg_policy}.
            
        \subsubsection*{Proof of Proposition \ref{P:policy_grad_th}:}
            In order to obtain the desired result, we exploit both the classical policy gradient theorem and the product derivative rule. 
            Using elementary calculus, we obtain the following: 
            \begin{align}
                \nabla_\theta \mathds{E}^{\eta}_{p_0,\pi_\theta}[Q_{\pi_\theta}^c] & = \nabla_\theta \int_{s_0, s_+,a_+}\hspace{-1cm} p_0(s_0) P^{\eta}_{\pi_\theta}(s_{+}, a_{+}|s_0) Q^c_{\pi_\theta}(s_+,a_+)  \\
                & = \nabla_\theta \int_{s_0, s_+,a_+}\hspace{-1cm} p_0(s_0) P^{\eta}_{\pi_\theta}(s_{+}|s_0) \pi_\theta(a_{+}|s_{+}) Q^c_{\pi_\theta}(s_+,a_+)\\
                & = \nabla_\theta \int_{s_0, s_+}\hspace{-0.6cm} p_0(s_0) P^{\eta}_{\pi_\theta}(s_{+}|s_0) v_{\pi_\theta}^c(s_{+})\\
                & = \int_{s_0, s_+}\hspace{-0.6cm} p_0(s_0) \Big[  v_{\pi_\theta}^c(s_{+}) \nabla_\theta P^{\eta}_{\pi_\theta}(s_{+}|s_0) + P^{\eta}_{\pi_\theta}(s_{+}|s_0)  \nabla_\theta v_{\pi_\theta}^c(s_{+}) \Big] \\
                & = \int_{s_0, s_+}\hspace{-0.6cm} p_0(s_0) v_{\pi_\theta}^c(s_{+}) \nabla_\theta P^{\eta}_{\pi_\theta}(s_{+}|s_0) + \underbrace{\int_{s_0, s_+}\hspace{-0.6cm} p_0(s_0) P^{\eta}_{\pi_\theta}(s_{+}|s_0) \nabla_\theta v_{\pi_\theta}^c(s_{+})}_A
            \end{align}
            Recall that the policy gradient theorem can be written simply as: 
            \begin{align}
                &\forall s_0\in\mathcal{S} \quad & \nabla_\theta \mathbb{E}_{\pi_\theta}[\sum_t \gamma^t c(s_t,a_t)|s_0] & = \nabla_\theta \int_a \pi_\theta(a|s_0) Q_{\pi_\theta}^c(s_0,a) \\
                & & & = \nabla_\theta v_{\pi_\theta}^c(s_0) = \int_{s} \rho_\pi(s|s_0) \mathbb{E}_{a\sim\pi_\theta}[Q_{\pi_\theta}^c(s,a) \nabla_\theta\log\pi_\theta(a|s)]
            \end{align}
            This concludes our proof as: 
            \begin{align}
                A = \int_{s_0, s_+}\hspace{-0.6cm} p_0(s_0) P^{\eta}_{\pi_\theta}(s_{+}|s_0) \nabla_\theta v_{\pi_\theta}^c(s_{+}) & = \int_{s_0, s_+, s}\hspace{-0.8cm} p_0(s_0) P^{\eta}_{\pi_\theta}(s_{+}|s_0) \rho_\pi(s|s_{+}) \mathbb{E}_{a\sim\pi_\theta}[Q_{\pi_\theta}^c(s,a) \nabla_\theta\log\pi_\theta(a|s)]
            \end{align}
            
        \subsubsection*{Proof of Proposition \ref{P:trpo_identity}:}
            The proof relies on the definition of the advantage function to obtain the first equality and on proposition \ref{tower_rule} to obtain the second one. Recall that:
            \begin{align}
                \mathbb{E}_{p_0,\pi_n}^\eta\Big[\sum_t \gamma^t A_{\pi_o}(s_t,a_t) \Big] & = \mathbb{E}_{p_0,\pi_n}^\eta\Big[\sum_t \gamma^t (c(s_t,a_t) + \gamma v_{\pi_o}^c(s_{t+1}) - v_{\pi_o}^c(s_t)) \Big]  \\
                & = \mathbb{E}_{p_0,\pi_n}^\eta[ - v_{\pi_o}^c(s_0) + \sum_t \gamma^t c(s_t,a_t) ] \\
                & = - \underbrace{\int_{s_0} p_0(s_0) v_{\pi_o}^c(s_0)}_{\mathcal{L}_0^\eta(\pi_o,c)} + \underbrace{\mathbb{E}_{p_0,\pi_n}^\eta[ \sum_t \gamma^t c(s_t,a_t) ]}_{\mathcal{L}_0^\eta(\pi_n,c)} \\
                \iff \mathcal{L}_0^\eta(\pi_n,c) & = \mathcal{L}_0^\eta(\pi_o,c) + \mathbb{E}_{p_0,\pi_n}^\eta\Big[\sum_t \gamma^t A_{\pi_o}(s_t,a_t) \Big]
            \end{align}
            This concludes the proof of the first equality. In addition, by observing that $A_\pi$ is a mapping from $\mathcal{S}\times\mathcal{A}$ to $\mathbb{R}$, we can apply proposition \ref{tower_rule} to further simplify the expectation term: 
            \begin{align}
                \mathbb{E}_{p_0,\pi_n}^\eta\Big[\sum_t \gamma^t A_{\pi_o}(s_t,a_t) \Big] = \int_{s_0,s,a,s_{+},a_{+}}\hspace{-1cm} p_0(s_0)\rho_\pi(s,a|s_0)P_\pi^\eta(s_{+},a_{+}|s,a) A_{\pi_o}(s_{+},a_{+})
            \end{align}
            This concludes the proof as we have: 
            \begin{align}
                \mathcal{L}_0^\eta(\pi_n,c) = \mathcal{L}_0^\eta(\pi_o,c) + \int_{s_0,s,a,s_{+},a_{+}}\hspace{-1cm} p_0(s_0)\rho_\pi(s,a|s_0)P_\pi^\eta(s_{+},a_{+}|s,a) A_{\pi_o}(s_{+},a_{+})
            \end{align}
    
    \subsection{Generalised Inverse Reinforcement Learning}
    \label{A:GIRL}
        In this section, we address the claims stated in section \ref{sec:GIRL}. 
        \subsubsection*{Proof of Proposition \ref{PR:equivalent}:}
            The proof relies on the properties of saddle point \cite{convex}. Let $\Tilde{c}$, $\Tilde{\pi}$ and $\hat{\pi}$ be respectively defined as: 
            \begin{align*}
                \Tilde{c} & \in \IRL_{\psi,\Omega}^\eta(\pi_E) \\
                \Tilde{\pi} & \in \RL_\Omega^\eta(\Tilde{c})=\RL_\Omega^\eta \circ \IRL_{\psi,\Omega}^\eta(\pi_E) = \argmin_\pi \mathcal{L}_\Omega^\eta(\pi,\Tilde{c}) \\
                \hat{\pi} & \in \argmin_\pi \max_c L(\pi,c)
            \end{align*}
            Our goal is to prove that $\Tilde{\pi}=\hat{\pi}$. Equivalently (due to proposition \ref{policy_occupancy}) this boils down to proving that $\mu_{\Tilde{\pi}}=\mu_{\hat{\pi}}$.
            Using Proposition \ref{tower_rule} and \ref{policy_occupancy}, we can re-write: 
            \begin{align}
                \mathcal{L}_\Omega^\eta(\pi,c) & := \mathds{E}^{\eta}_{p_0,\pi}\big[ \sum_t \gamma^t c(s_t,a_t) \big] - \Omega(\pi) \\
                & = \int_{s_0,s,a,s_{+},a_{+}}\hspace{-1cm} p_0(s_0)\rho_\pi(s,a|s_0)P_\pi^\eta(s_{+},a_{+}|s,a) c(s_{+},a_{+}) - \Omega(\pi)  \\
                & = \Bar{\mathcal{L}}_\Omega^\eta(\mu_\pi,c) = \int_{s_0,s_{+},a_{+}}\hspace{-.5cm} p_0(s_0)\mu_\pi(s_{+},a_{+}|s_0) c(s_{+},a_{+}) - \Omega(\mu_\pi)
            \end{align}
            This implies that :
            \begin{align}
                \mu_{\Tilde{\pi}} & \in \argmin_{\mu\in\mathcal{D}} \Bar{\mathcal{L}}_\Omega^\eta(\mu,\Tilde{c}) =  \argmin_{\mu\in\mathcal{D}} \Bar{L}(\mu, \Tilde{c}) 
                \label{RES_0}
            \end{align}
            where:
            \begin{align}
                \Bar{L}& : \mathcal{D}\times\mathbb{R}^{\mathcal{S}\times\mathcal{A}} \rightarrow \mathbb{R}\\
                \mathcal{D} & = \Big\{ \mu_\pi: \mu_\pi(s,a|s_0) = \sum_{t,k} \gamma^t \eta(k) \mathbb{P}_\pi(s_{t+k}=s,a_{t+k}=a|s_0) \Big| \pi:\mathcal{S}\rightarrow\Delta(\mathcal{A})\Big\} \\ 
                \Bar{L}(\mu_\pi, c) & = L(\pi, c) = -\Omega(\pi) - \psi(c) + \int_{s_0} p_0(s_0) d_c(\pi\|\pi_E)(s_0) \\
                & = -\Omega(\mu_\pi) - \psi(c) + \int_{s_0,s_{+},a_{+}}\hspace{-.5cm} p_0(s_0) c(s_{+},a_{+}) \Big[ \mu_\pi(s_{+},a_{+}|s_0) - \mu_{\pi_E}(s_{+},a_{+}|s_0) \Big]
            \end{align}
            In addition, for the same reasons, we have that: 
            \begin{align}
            \begin{split}
                \Tilde{c} & \in \argmax_c \min_{\pi} \mathcal{L}_\Omega^\eta(\pi,c) -  \mathcal{L}_\Omega^\eta(\pi_E^*,c)   - \psi(c)  = \argmax_c \min_{\mu\in\mathcal{D}} \Bar{L}(\mu, c) \\
                \mu_{\hat{\pi}} & \in \argmin_{\mu\in\mathcal{D}} \max_c \Bar{L}(\mu, c)
            \end{split}
            \label{RES_1}
            \end{align}
            Notice that $\mathbb{R}^{\mathcal{S}\times\mathcal{A}}$ is convex, $\Bar{L}(\mu, c)$ is convex w.r.t $\mu$ and concave w.r.t $c$ (due to convexity of $\psi$ and $-\Omega$). Therefore, the minmax duality property holds as soon as $\mathcal{D}$ is compact and convex. 
            
            \paragraph{Convexity and compacity of $\mathcal{D}$:}
            We prove this under the assumption that $\eta$ is a geometric distribution (i.e. $\eta=\Geom(\gamma_\eta)$). To establish the convexity, we prove that $\mathcal{D}=\{f:\mathcal{S}\rightarrow\Delta(\mathcal{S}\times\mathcal{A})\}$ where $f$ is a solution of the following equations:
            \begin{align}
            &\forall s,s_0\in\mathcal{S}, a\in\mathcal{A} \quad & 
            \begin{split}
                & \int_{a} f(s,a|s_0) = \int_{a} g(s,a|s_0) + \int_{s',a'} \gamma f(s',a'|s_0) \mathcal{P}(s|s',a')  \\
                & \int_{a} f(s,a|s_0) = \int_{a} h(s,a|s_0) + \int_{s',a'} \gamma_\eta f(s',a'|s_0) \mathcal{P}(s|s',a')  \\
                & \int_{a} g(s,a|s_0) = 1 + \int_{s',a'} \gamma_\eta \mathcal{P}(s|s',a') g(s',a'|s_0)\\
                & \int_{a} h(s,a|s_0) = 1 + \int_{s',a'} \gamma \mathcal{P}(s|s',a') h(s',a'|s_0)\\
                & f(s,a|s_0)\geq0, \quad  g(s,a|s_0)\geq0, \quad  h(s,a|s_0)\geq0
            \end{split}. &
            \label{feasible_set_equivalent}
            \end{align}
            To this end, notice that for any policy $\pi$, we can verify that $(f=\mu_\pi, g=P_\pi^\eta, h=\rho_\pi)$ is a solution to Equation \ref{feasible_set_equivalent}. We focus now on the converse statement. Let~$(f,g,h)$ a solution to Equation \ref{feasible_set_equivalent}. Given the third equality: 
            \begin{align}
                \int_{a} g(s,a|s_0) = 1 + \int_{s',a'} \gamma_\eta \mathcal{P}(s|s',a') g(s',a'|s_0),
            \end{align}
            we exploit a classical result from the MDP literature [\cite{puterman2014markov}, Section~6.9.2], to derive the existence of a policy $\pi_g$ such that:
            \begin{align}
            \begin{split}
                & g(s,a|s_0) = \pi_g(a|s) \Big[ 1 + \int_{s',a'} \gamma_\eta \mathcal{P}(s|s',a') g(s',a'|s_0) \Big] \\
                \textbf{where } \forall s_0 \in \mathcal{S} \quad & \pi_g(a|s) = \frac{g(s,a|s_0)}{\int_{a'} g(s,a'|s_0)}
            \end{split}
            \label{P_feasible_set}
            \end{align}
            From Equation \ref{P_feasible_set}, we conclude that $g$ is the unique fixed point of a $\gamma_\eta$-contraction. We can verify that $g=P_{\pi_g}^\eta$ is the unique solution of Equation \ref{P_feasible_set}. Using a similar reasoning with respect to the fourth equality in Equation \ref{feasible_set_equivalent}, we conclude that $h=\rho_{\pi_h}$ where for any state $s_0\in\mathcal{S}$, we have $\pi_h(a|s) = \frac{h(s,a|s_0)}{\int_{a'} h(s,a'|s_0)}$. From this, we conclude that for any solution $(f,g,h)$ to Equation \ref{feasible_set_equivalent}, there exist two policies $\pi_h$ and $\pi_g$ such that:
            \begin{align}
            &\forall s,s_0\in\mathcal{S}, a\in\mathcal{A} \quad & 
            \begin{split}
                & \int_{a} f(s,a|s_0) = P_{\pi_g}^\eta(s|s_0) + \int_{s',a'} \gamma f(s',a'|s_0) \mathcal{P}(s|s',a')  \\
                & \int_{a} f(s,a|s_0) = \rho_{\pi_h}(s|s_0) + \int_{s',a'} \gamma_\eta f(s',a'|s_0) \mathcal{P}(s|s',a')  \\
                & f(s,a|s_0)\geq0
            \end{split}. &
            \label{feasible_set_equivalent_simpler_integrated}
            \end{align}
            which is equivalent to [\cite{puterman2014markov}, Section~6.9.2]: 
            \begin{align}
            &\forall s,s_0\in\mathcal{S}, a\in\mathcal{A} \quad & 
            \begin{split}
                & f(s,a|s_0) = \pi_f(a|s) \Big[ P_{\pi_g}^\eta(s|s_0) + \int_{s',a'} \gamma f(s',a'|s_0) \mathcal{P}(s|s',a') \Big] \\
                & f(s,a|s_0) = \pi_f(a|s) \Big[ \rho_{\pi_h}(s|s_0) + \int_{s',a'} \gamma_\eta f(s',a'|s_0) \mathcal{P}(s|s',a') \Big]  \\
                & \pi_f(a|s) = \frac{f(s,a|s_0)}{\int_{a'} f(s,a'|s_0)}, \quad f(s,a|s_0)\geq0
            \end{split}. & 
            \label{feasible_set_equivalent_simpler}
            \end{align}
            
            Notice that the first equality in Equation \ref{feasible_set_equivalent_simpler}, implies that $f$ is the unique fixed point of a $\gamma$-contraction. We also notice that:
            \begin{align}
                f_{\pi_g, \pi_f}^0(s_+, a_+ | s_0) & := \sum_{k,t} \gamma^t \gamma_\eta^k \int_{s,a} \mathbb{P}_{\pi_f}^t(s_+,a_+|s) \mathbb{P}_{\pi_g}^k(s,a|s_0) \\
                & = \sum_{k,t>0} \gamma^t \gamma_\eta^k \int_{s,a} \mathbb{P}_{\pi_f}^t(s_+,a_+|s) \mathbb{P}_{\pi_g}^k(s,a|s_0) + \sum_k  \gamma_\eta^k \int_{s,a} \mathbb{P}_{\pi_f}^0(s_+,a_+|s) \mathbb{P}_{\pi_g}^k(s,a|s_0) \\
                & = \gamma\pi_f(a_+|s_+)\int_{s',a'}  f_{\pi_g, \pi_f}^0(s',a'|s_0) \mathcal{P}(s|s',a') + \sum_k  \gamma_\eta^k \mathbb{P}_{\pi_g}^k(s_+|s_0) \pi_f(a_+|s_+) \\
                & = \pi_f(a_+|s_+) \Big[ \int_{s',a'} \gamma f_{\pi_g, \pi_f}^0(s',a'|s_0) \mathcal{P}(s_+|s',a') + P_{\pi_g}^\eta(s_+|s_0)  \Big]
            \end{align}
            where $\mathbb{P}_\pi^n(s,a|s^0) = \mathbb{P}_\pi(s_n=s, a_n=a|s_0=s^0)$.
            Thus, we conclude $f_{\pi_g, \pi_f}^0$ is the unique solution to the first equality. Similarly, we notice that the second equality is a $\gamma_\eta$-contraction, whose unique fixed point is: 
            \begin{align}
                f_{\pi_h, \pi_f}^1(s_+, a_+ | s_0) & := \sum_{k,t} \gamma^t \gamma_\eta^k \int_{s,a} \mathbb{P}_{\pi_f}^k(s_+,a_+|s) \mathbb{P}_{\pi_h}^t(s,a|s_0) \\
                & = \sum_{k>0,t} \gamma^t \gamma_\eta^k \int_{s,a} \mathbb{P}_{\pi_f}^k(s_+,a_+|s) \mathbb{P}_{\pi_h}^t(s,a|s_0) + \sum_t  \gamma^t \int_{s,a} \mathbb{P}_{\pi_f}^0(s_+,a_+|s) \mathbb{P}_{\pi_h}^t(s,a|s_0) \\
                & = \pi_f(a_+|s_+) \Big[ \int_{s',a'} \gamma_\eta f_{\pi_h, \pi_f}^1(s',a'|s_0) \mathcal{P}(s_+|s',a') + \rho_{\pi_h}(s_+|s_0) \Big]
            \end{align}
            
            We derive from the previously discussed statement, that if $(f,g,h)$ is a solution to Equation \ref{feasible_set_equivalent}, then there exist three policies $(\pi_f,\pi_g, \pi_h)$ such that: 
            \begin{align}
            \begin{split}
                f & = f_{\pi_g, \pi_f}^0 = f_{\pi_h, \pi_f}^1 \\
                g & = P_{\pi_g}^\eta \\
                h & = \rho_{\pi_h}
            \end{split}
            \label{policy_condition}
            \end{align}
            
            However, not any random choice of policies $(\pi_f,\pi_g, \pi_h)$ can satisfy Equation \ref{policy_condition}. By varying $\gamma$ and $\gamma_\eta$, we notice that in order for the first equality to hold, the following equality must be satisfied for any integers $(k,t)$, and for any states $(s_+,a_+,s_0)$:
            \begin{align}
                \int_{s,a}\mathbb{P}_{\pi_f}^t(s_+,a_+|s) \mathbb{P}_{\pi_g}^k(s,a|s_0) = \int_{s,a}\mathbb{P}_{\pi_f}^k(s_+,a_+|s) \mathbb{P}_{\pi_h}^t(s,a|s_0)
            \end{align}
            by fixing $k$ at zero and varying $t$ and by fixing $t$ at zero and varying $k$, we obtain the following constraints: 
            \begin{align}
                \begin{split}
                    P_{\pi_f}^\eta & = P_{\pi_g}^\eta \\
                    \rho_{\pi_f} & = \rho_{\pi_h}
                \end{split}
                \label{policy_equality}
            \end{align}
            Using Proposition \ref{policy_occupancy} and Equation \ref{policy_equality}, it follows that Equation \ref{policy_condition} admits a solution if and only if $\pi_f=\pi_g=\pi_h$. This means that if $(f,g,h)$ is a solution to Equation \ref{feasible_set_equivalent}, then there exists a policy $\pi$ such that: 
            \begin{align}
            \begin{split}
                f & = f_{\pi, \pi}^0 = f_{\pi, \pi}^1 = \mu_\pi \\
                g & = P_{\pi}^\eta \\
                h & = \rho_{\pi}
            \end{split}
            \end{align}
            This concludes the converse statement, proving that $\mathcal{D}$ is a set of occupancy measures satisfying the set of affine constraints from Equation \ref{feasible_set_equivalent}. Consequently, $\mathcal{D}$ is a convex set. 
            In addition, the limit of any sequence of elements from $\mathcal{D}$ will also satisfy Equation \ref{feasible_set_equivalent_simpler_integrated}. From this we establish that $\mathcal{D}$ is closed which implies that it is also compact.
            
            From this we derive that minmax duality holds and that:
            \begin{align}
                \min_{\mu\in\mathcal{D}} \max_c \Bar{L}(\mu, c) = \max_c \min_{\mu\in\mathcal{D}} \Bar{L}(\mu, c)
            \end{align}
            From Equation \ref{RES_1}, it follows that $(\mu_{\hat{\pi}}, \Tilde{c})$ is a saddle point for the function $\Bar{L}$. This implies from Equation \ref{RES_0} that:
            \begin{align}
                \mu_{\hat{\pi}}, \mu_{\Tilde{\pi}} \in \argmin_{\mu\in\mathcal{D}} \Bar{L}(\mu, \Tilde{c}) 
            \end{align}
            In addition, due to the strict convexity of $\Bar{L}$ w.r.t $\mu$ (due to assumed strict convexity of $\Omega$) we have that: 
            \begin{align}
                \mu_{\hat{\pi}} = \mu_{\Tilde{\pi}}
            \end{align}
            which concludes our proof.
        
        \subsubsection*{Proof of Corollary \ref{CR:Nash}:}
            The proof entails directly from the duality of $\Bar{L}$ and that $(\Tilde{c}, \mu_{\Tilde{\pi}})$ is a saddle point of $\Bar{L}$. 
        
        \subsubsection*{Proof of Proposition \ref{P:convex_entropy}:}
            The proof relies on re-writing the $\eta$-weighted entropy reguliser using Proposition \ref{tower_rule}, and then verifying its convexity with respect to $\mu_\pi$ using the log-sum inequality. 
            In fact, notice that: 
            \begin{align}
                H^{\eta}_{p_0}(\pi) & = \mathds{E}^{\eta}_{p_0,\pi}\Big[\sum_t -\gamma^t\log\big[\pi(a_t|s_t)\big]\Big] \\
                & = \int_{s_0,s_{+},a_{+}}\hspace{-.5cm} p_0(s_0)\mu_\pi(s_{+},a_{+}|s_0) \log\big[\pi(a_{+}|s_{+})\big] = \Bar{H}^{\eta}_{p_0}(\mu_\pi)
            \end{align}
            Consider two $\eta$-weighted occupancy measures $\mu_1,\mu_2$, and let $\pi_1,\pi_2$ their respective policies. Let $\lambda\in]0,1[$:
            \begin{align}
                \Bar{H}(\lambda \mu_1 + (1-\lambda)\mu_2) & = \int_{s_0,s_{+},a_{+}}\hspace{-.7cm} -p_0(s_0)\big[\lambda \mu_1 + (1-\lambda)\mu_2\big](s_{+},a_{+}|s_0) \log\Big[\frac{\big[\lambda \mu_1 + (1-\lambda)\mu_2\big](s_{+},a_{+}|s_0)}{\int_{a}\big[\lambda \mu_1 + (1-\lambda)\mu_2\big](s_{+},a|s_0)}\Big]
            \end{align}
            Du to the log-sum inequality we have:
            \begin{align}
                & \big[\lambda \mu_1 + (1-\lambda)\mu_2\big](s_{+},a_{+}|s_0) \log\Big[\frac{\big[\lambda \mu_1 + (1-\lambda)\mu_2\big](s_{+},a_{+}|s_0)}{\int_{a}\big[\lambda \mu_1 + (1-\lambda)\mu_2\big](s_{+},a|s_0)}\Big] \\
                = & \big[\lambda \mu_1 + (1-\lambda)\mu_2\big](s_{+},a_{+}|s_0) \log\Big[\frac{\lambda\mu_1(s_{+},a_{+}|s_0) + (1-\lambda)\mu_2(s_{+},a_{+}|s_0) }{\lambda\int_{a}\mu_1(s_{+},a|s_0) + (1-\lambda)\int_{a}\mu_2(s_{+},a|s_0)}\Big] \\
                \leq & \lambda \mu_1 \log\Big[\frac{\lambda\mu_1(s_{+},a_{+}|s_0)}{\lambda\int_{a}\mu_1(s_{+},a|s_0)}\Big] + (1-\lambda) \mu_2 \log\Big[\frac{(1-\lambda)\mu_2(s_{+},a_{+}|s_0)}{(1-\lambda)\int_{a}\mu_2(s_{+},a|s_0)}\Big] \\
                = & \lambda \mu_1 \log\Big[\frac{\mu_1(s_{+},a_{+}|s_0)}{\int_{a}\mu_1(s_{+},a|s_0)}\Big] + (1-\lambda) \mu_2 \log\Big[\frac{\mu_2(s_{+},a_{+}|s_0)}{\int_{a}\mu_2(s_{+},a|s_0)}\Big] 
            \end{align}
            This implies that:
            \begin{align}
                \Bar{H}(\lambda \mu_1 + (1-\lambda)\mu_2) \geq \lambda \Bar{H}(\mu_1) + (1-\lambda) \Bar{H}(\mu_2)
            \end{align}
            with equality if and only if $\pi_1(a|s) := \frac{\mu_1(s,a|s_0)}{\int_{a'}\mu_1(s,a'|s_0)} =  \frac{\mu_2(s,a|s_0)}{\int_{a'}\mu_2(s,a'|s_0)} := \pi_2(a|s)$. This concludes the proof of the $\eta$-weighted strict concavity w.r.t the set of measures $\mu$.
            
    \subsection{Data augmentation}
    \label{A:Idle_data_augmentation}
        In this section, we provide the proof of Proposition \ref{P:Idle}. 
        We start by noticing that $V(D,G)$ is the loss function used by conditional generative adversarial neural networks \cite{CGAN}, which minimum w.r.t the discriminator is achieved for the optimal Bayes classifier \cite{GAN}:
        \begin{align}
            D^*(s,a|s_0) = \frac{P_\pi^\eta(s,a|s_0)}{P_\pi^\eta(s,a|s_0) + G(s,a|s_0)}
        \end{align}
        where $G(s,a|s_0)$ is the probability of generating $(s,a)$ using the generator $G$. From this, we can re-write the generator's loss against an infinite capacity (optimal) discriminator as:
        \begin{align}
            V(D^*,G) & = D_{KL}(P_\pi^\eta(s,a|s_0) \| \frac{P_\pi^\eta(s,a|s_0)}{P_\pi^\eta(s,a|s_0) + G(s,a|s_0)}) + D_{KL}(G(s,a|s_0) \| \frac{P_\pi^\eta(s,a|s_0)}{P_\pi^\eta(s,a|s_0) + G(s,a|s_0)}) - \log(4) \\
            & = 2D_{JSC}(G(s,a|s_0) \| P_\pi^\eta(s,a|s_0)) - \log(4)
        \end{align}
        where $D_{KL}$ is the KL divergence, and $D_{JSC}$ is the Jenson-Shannon divergence. A global minimum is achieved when $G^*$:
        \begin{align}
            G^*(s,a|s_0)=P_\pi^\eta(s,a|s_0).
        \end{align}
        This concludes the proof as it implies that $(\Tilde{D}=\frac{1}{2},\Tilde{G}=P_{\pi}^{\eta})$ is a Nash-equilibrium
        
    \subsection{Particular settings of interest}
    \label{A:particular_settings}
        In this section we address the claims stated in section \ref{sec:settings}.
        \subsubsection*{Proof of Proposition \ref{P:EMMA}:}
            Notice that in this setting: 
            \begin{align}
                \max_{c} L(\pi, c) = & \max_{c\in\mathcal{C}_\textit{linear}} \int_{s_0} p_0(s_0) d_c(\pi\|\pi_E)(s_0)  \\
                = & \max_{c\in\mathcal{C}_\textit{linear}} \int_{s_0} p_0(s_0) \big[\mu_\pi(s_{+},a_{+}|s_0)-\mu_{\pi_E}(s_{+},a_{+}|s_0)\big] c(s_{+},a_{+}) \\
                = & \max_{w \textbf{ with } \|w\|_2\leq 1} \int_{s_0} p_0(s_0) \big[\mu_\pi(s_{+},a_{+}|s_0)-\mu_{\pi_E}(s_{+},a_{+}|s_0)\big] \sum_i w_i f_i(s_{+},a_{+}) \\
                = & \max_{w \textbf{ with } \|w\|_2\leq 1} \sum_i w_i \int_{s_0} p_0(s_0) \big[\mu_\pi(s_{+},a_{+}|s_0)-\mu_{\pi_E}(s_{+},a_{+}|s_0)\big]  f_i(s_{+},a_{+}) \\
                = & \Big\| \int_{s_0} p_0(s_0) \big[\mu_\pi(s_{+},a_{+}|s_0)-\mu_{\pi_E}(s_{+},a_{+}|s_0)\big]  f_i(s_{+},a_{+}) \Big\|_2 \\
                = & \Big\|\mathbb{E}_{\mu_\pi}[f]-\mathbb{E}_{\mu_{\pi_E}}[f] \Big\|_2
            \end{align}
            This concludes the proof.
            
        \subsubsection*{Proof of Proposition \ref{P:WIEM}:}
            In this case, we notice the following:
            \begin{align}
                \max_{c} L(\pi, c) = & \max_{c\in\mathcal{C}_\textit{convex}} \int_{s_0} p_0(s_0) d_c(\pi\|\pi_E)(s_0)  \\
                = & \max_{c\in\mathcal{C}_\textit{convex}} \int_{s_0} p_0(s_0) \big[\mu_\pi(s_{+},a_{+}|s_0)-\mu_{\pi_E}(s_{+},a_{+}|s_0)\big] c(s_{+},a_{+}) \\
                = & \max_{w_i>0 \textbf{ with } \sum w_i = 1} \int_{s_0} p_0(s_0) \big[\mu_\pi(s_{+},a_{+}|s_0)-\mu_{\pi_E}(s_{+},a_{+}|s_0)\big] \sum_i w_i f_i(s_{+},a_{+}) \\
                = & \max_{w_i>0 \textbf{ with } \sum w_i = 1} \sum_i w_i \int_{s_0} p_0(s_0) \big[\mu_\pi(s_{+},a_{+}|s_0)-\mu_{\pi_E}(s_{+},a_{+}|s_0)\big]  f_i(s_{+},a_{+}) \\
                = & \max_i \int_{s_0} p_0(s_0) \big[\mu_\pi(s_{+},a_{+}|s_0)-\mu_{\pi_E}(s_{+},a_{+}|s_0)\big]  f_i(s_{+},a_{+})\\
                = & \max_i \mathbb{E}_{\mu_\pi}[f_i]-\mathbb{E}_{\mu_{\pi_E}}[f_i] 
            \end{align}
            This concludes the proof.
            
        \subsubsection*{Proof of Proposition \ref{P:MEGAN}:}
            We start by re-writing the cost function as an expectation with respect to the occupancy measure $\mu_\pi$ using Proposition \ref{tower_rule}: 
            \begin{equation}
                \psi_{GAN}(c) = 
                \left\{
                \begin{array}{ll}
                    \int_{s,a,s_0} p_0(s_0) \mu_{\pi_E}(s,a|s_0) g(c(s,a)) & \textit{ if } c<0 \\
                    +\infty &  \textit{ otherwise }
                \end{array}
                \right .
            \end{equation}
            With this, we can rewrite the objective function $L(\pi,c)$ in this setting as follows:
            \begin{align}
                L(\pi, c) & = -\Omega(\pi) - \psi(c) + \int_{s_0} p_0(s_0) d_c(\pi\|\pi_E)(s_0)\\
                & = -\Omega(\pi) + \int_{s_0} p_0(s_0) \int_{s,a} \Big[ \mu_{\pi_E}(s,a|s_0) g(c(s,a)) + \big( \mu_{\pi}(s,a|s_0) - \mu_{\pi_E}(s,a|s_0)  \big) c(s,a) \Big]
            \end{align}
            Notice that this is the same objective as the one used in GAIL [\cite{gail} Appendix A.2], where we compute expectations with respect to $\mu_\pi$ while the do it with respect to $\rho_\pi$.
            Using the same change of variable, we can obtain the folowing: 
            \begin{align}
                \max_c L(\pi, c) & = \max_{D\in(0,1)^{\mathcal{S}\times\mathcal{A}}} -\Omega(\pi) + \int_{s_0} p_0(s_0) \int_{s,a} \mu_{\pi}(s,a|s_0) \log(D(s,a) - \mu_{\pi_E}(s,a|s_0) \log(1-D(s,a))
            \end{align}
            
            This concludes the proof of Proposition \ref{P:MEGAN}, as we can state using Proposition \ref{tower_rule}:
            \begin{align}
                \RL_\Omega^\eta \circ \IRL_{\psi}^\eta (\pi_E) & = \argmin_\pi \max_c L(\pi,c) \\
                & = \argmin_\pi -\Omega(\pi) + \max_{D \in (0,1)^{\mathcal{S}\times\mathcal{A}}} \mathbb{E}_{p_0,\pi}^\eta[\log D]-\mathbb{E}_{p_0,\pi_E}^\eta[\log (1-D))] 
            \end{align}
%\input{Appendix/prop2}
%\input{Appendix/off policy data augmentation} % USEFUL FOR OFF-LINE IRL (future work)

%%%%%%%%%%%%%%%%%%%%%%%%%%%%%%%%%%%%%%%%%%%%%%%%%%%%%%%%%%%%%%%%%%%%%%%%%%%%%%%

\end{document}